\documentclass[sn-mathphys-num]{sn-jnl}

\usepackage{amsmath,amsfonts}
\usepackage{amssymb}
\usepackage{amsthm}
\usepackage{algorithm}
\usepackage{array}
\usepackage[caption=false,font=normalsize,labelfont=sf,textfont=sf]{subfig}
\usepackage{textcomp}
\usepackage{stfloats}
\usepackage{url}
\usepackage{verbatim}
\usepackage{graphicx}
\usepackage{cite}
\usepackage{multirow}
\usepackage{booktabs}
\usepackage{pifont}

\usepackage{mathrsfs}
\usepackage{xcolor}
\usepackage{manyfoot}
\usepackage{algorithmicx}
\usepackage{algpseudocode}
\usepackage{listings}
\usepackage{color}

\theoremstyle{thmstyleone}%
\theoremstyle{thmstyletwo}%

\theoremstyle{thmstylethree}%

\newcommand{\xmark}{\ding{55}}
\begin{document}

\title[Live Video Captioning]{Live Video Captioning}


\author[1]{\fnm{Eduardo} \sur{Blanco-Fern\'andez}}\email{e.blancof@edu.uah.es}

\author[1]{\fnm{Carlos} \sur{Guti\'errez-\'Alvarez}}\email{carlos.gutierrezalva@uah.es}

\author[2]{\fnm{Nadia} \sur{Nasri}}\email{nadia.nasri@ua.es}

\author[1]{\fnm{Saturnino} \sur{Maldonado-Basc\'on}}\email{saturnino.maldonado@uah.es}

\author*[1]{\fnm{Roberto~J.} \sur{L\'opez-Sastre}}\email{robertoj.lopez@uah.es}

\affil*[1]{\orgdiv{Department of Signal Theory and Communications}, \orgname{University of Alcal\'a}, \orgaddress{\city{Alcal\'a de Henares}, \postcode{28805}, \country{Spain}}}

\affil[2]{\orgdiv{Department of Computer Science and Artificial Intelligence}, \orgname{University of Alicante}, \orgaddress{\city{San Vicente del Raspeig}, \postcode{03690}, \country{Spain}}}

\abstract{
Dense video captioning involves detecting and describing events within video sequences.
Traditional methods operate in an offline setting, assuming the entire video is available for analysis.
In contrast, in this work we introduce a groundbreaking paradigm: Live Video Captioning (LVC), where captions must be generated for video streams in an online manner.
This shift brings unique challenges, including processing partial observations of the events and the need for a temporal anticipation of the actions.
\\
We formally define the novel problem of LVC and propose innovative evaluation metrics specifically designed for this online scenario, demonstrating their advantages over traditional metrics.
To address the novel complexities of LVC, we present a new model that combines deformable transformers with temporal filtering, enabling effective captioning over video streams.
\\
Extensive experiments on the ActivityNet Captions dataset validate the proposed approach, showcasing its superior performance in the LVC setting compared to state-of-the-art offline methods.
To foster further research, we provide the results of our model and an evaluation toolkit with the new metrics integrated at: \url{https://github.com/gramuah/lvc}.
}


\maketitle

\section{Introduction}
As a growing field within video understanding, video captioning has gathered significant attention recently~\citep{Vaishnavi2024,Babavalian2024,Liu2024,Yao2024,Tang2023}.
The goal of these video captioning models is to produce a natural sentence that encapsulates the primary event in a \emph{short} video.
These models utilize datasets tailored to the described problem (e.g., MSR-VTT~\citep{xu2016-msr-vtt}, VATEX~\citep{Wang2019vatex}), where short video segments and their corresponding annotations in the form of captions are provided.
Nevertheless, because real-world videos are often lengthy, untrimmed, and feature multiple simultaneous events alongside background content, the aforementioned single-sentence video captioning models typically produce lackluster and less informative descriptions.
To tackle this more complex scenario, captioning approaches must both locate and describe the events occurring in long videos; this problem is known as \emph{dense} video captioning~\citep{Vaishnavi2024,Artham2024,Yang2023,wang2021,show_tell_sum,wang2020event}.

The majority of real-life videos encompass numerous events that may unfold simultaneously.
For instance, in a video featuring ``a waiter carrying food to a table'' one may also observe another ``individual eating and drinking'' or a ``woman sitting down''.
Dense video captioning models must generate descriptions for each event unfolding in a video, precisely indicating the start and end times of each event.

\begin{figure}[t]
    \centering
    \includegraphics[width=0.5\linewidth]{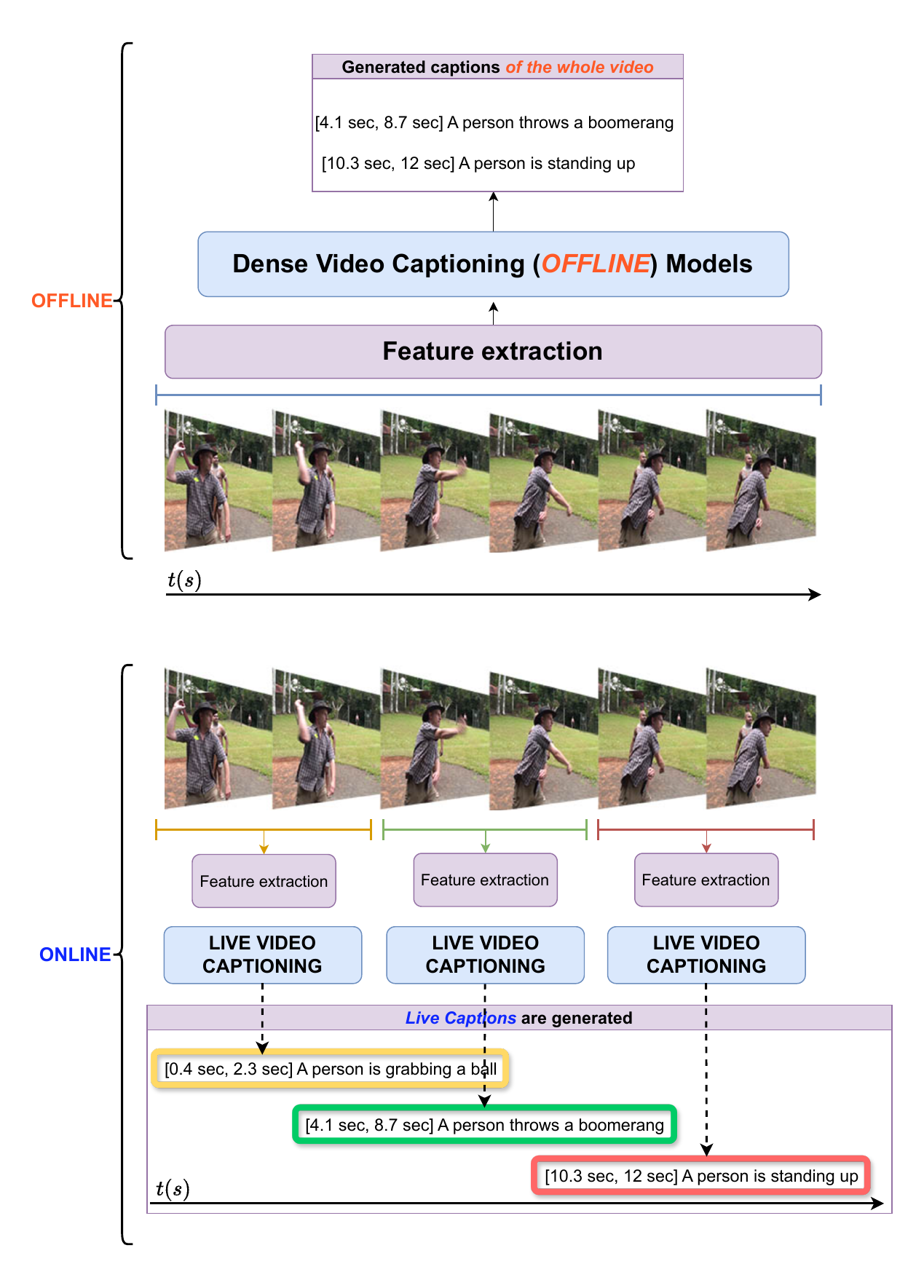}
    \caption{Above: Traditional models of dense video captioning work offline, accessing the whole video to generate
    the captions. Down: The live video captioning models must generate the captions for the video stream, in an
    online manner, and working with partial observations of the video.}
    \label{fig:lvc_idea}
\end{figure}


However, the current state-of-the-art in dense video captioning primarily offers \emph{offline} solutions.
As illustrated in the top part of Figure~\ref{fig:lvc_idea}, these models assume access to the entire video before generating captions.
In contrast, our work introduces a paradigm shift termed Live Video Captioning (LVC), depicted in the bottom part of Figure~\ref{fig:lvc_idea}.
In LVC, the challenge is to generate dense captions operating in an \emph{online} manner.
This introduces significant new challenges for traditional offline models.
Firstly, in an LVC scenario, it is not feasible to access the entire video to generate dense captions.
LVC models must work with \emph{partial} video observations, ideally through video streams, continuously generating dense captions as events unfold.
This restriction is particularly detrimental to models requiring an action proposal generation phase, e.g.~\citep{krishna2017dense,show_tell_sum,zhou2018}.
Secondly, LVC models must be capable of anticipating actions.
Working with partial observations limits the semantic information available compared to traditional offline models, making event identification more challenging.
Lastly, LVC solutions need to incorporate temporal attention and filtering mechanisms to refine their caption predictions as the video streams evolve.

From an applicability standpoint, the LVC problem introduces new scenarios where live captioning is crucial, and traditional offline models would fail to perform effectively.
LVC solutions must strike a balance between precision and speed, as both are equally vital.
The utility of LVC models is substantial, enabling a wide range of applications, from content summarization and accessibility support in live broadcasts to online perception and decision-making in robotics and video surveillance.
While the former paves the way for novel, accessible, and interactive multimedia experiences, the latter will be crucial for the next generation of autonomous navigation and human-robot interaction systems.
These are, without question, scenarios where traditional offline video captioning models fall short.

Interestingly, to the best of our knowledge, the problem of Live Video Captioning (LVC) has not been extensively investigated.
In this work, we make several key contributions.
First, we offer a formal description of the LVC problem, detailed in Section~\ref{sec:lvc_formulation}.
Second, we introduce a novel model for LVC that combines deformable transformers with a temporal filtering mechanism to generate dense captions over video streams, as described in Section~\ref{sec:lvcmodel}.
Third, we propose new evaluation metrics specifically designed for the online scenario, demonstrating that traditional offline metrics are inadequate, as discussed in Section~\ref{sec:novel_metric}.
Finally, in Section~\ref{sec:experiments}, we provide a comprehensive experimental evaluation using the ActivityNet Captions dataset, showcasing both the performance of our new LVC model and the effectiveness of the new metrics.
We also include a comparison with state-of-the-art offline methods, highlighting their limitations in addressing the new online scenario

\section{Related work}
\label{sec:related_work}
\textbf{Dense video captioning} presents a multifaceted challenge, intertwining event localization and event captioning.
Krishna et al.~\citep{krishna2017dense} introduced the inaugural dense video captioning model, incorporating a multi-scale proposal module for localization alongside an attention-based LSTM for contextually informed caption generation.
Subsequent research endeavors have aimed to enhance event representations through various means, including context modeling~\citep{wang2018bidirectional,yang2018hierarchical}, event-level relationships~\citep{wang2020event}, and multi-modal feature fusion~\citep{Iashin2020MDVC,Yang2023}, thus facilitating more precise and informative caption generation.

Previous methods have struggled with integrating the localization and captioning modules effectively.
Attempts to address this issue have led researchers to explore interactions between the two sub-tasks. Li et al.~\citep{li2018jointly} propose a proxy task, predicting language rewards of generated sentences, to enhance the optimization of the localization module.
Similarly, Zhou et al.~\citep{zhou2018end} introduce a differential masking mechanism, linking the gradient flow from captioning loss to proposals' boundaries, thereby facilitating joint optimization.
In \citep{wang2021}, the proposed approach exploits inter-task interactions by ensuring both sub-tasks share the same intermediate features.
Furthermore, the method employs a one-to-one matching strategy between intermediate feature vectors and target event instances, resulting in discriminative features for captioning.

All the aforementioned methods share an important feature: they tackle the problem using pipelines designed to operate \emph{offline}.
In other words, the results of all these models are optimal when they have access to the entire video for which they generate the dense captions.

We propose in this work an approach that addresses the problem of dense video captioning in an \emph{online} fashion.
This new problem is named as live video captioning (LVC).
Ideally, in the LVC problem, the captions must be generated as soon as possible, by processing the video stream.
This means that the models need to be adapted to work with partial observations of the video content, and, under this condition, produce dense captions as accurate as possible.
Note that other online approaches have been explored, for example, in the different problem of action detection (e.~g. \citep{DeGeest2016,Gao2017c,DeGeest2018,li2016,baptista-rios2020,lopez-sastre2021,hu2022}).

For the dense video captioning problem, only, to the best of our knowledge, Hori et al.~\citep{Hori2021} have proposed a multi-modal captioning approach that uses a timing detector so that the captions can be generated in the early stages of an event-triggered video clip.
This problem can be termed as early video captioning, where the target consists in evaluating the latency ratio needed to reach the same performance of an offline video captioning model for an event-triggered video.
Similar simplified experimental setups where explored in the context of early event detection in video, e.g.~\citep{Hoai2012,Huang2014,Kong2014}.
Note that these problems are different from our live video captioning.
We claim these simplified setups are not representative for practical applications, where occurrences of possibly many different actions need to be detected and a correct caption generated in an online manner, in long video recordings with widely varying content.
When it comes to a live video captioning system, the model should be continuously processing the video stream, and,
when necessary, producing dense video captions.
This necessitates precisely recognizing the current action at any point of its development.
Furthermore, to complete the LVC task, one must distinguish the action from a range of negative inputs, such as
background frames in which no pertinent actions are occurring.

Overall, we propose a live video captioning model that parallelizes localization, selection, and captioning tasks within a single end-to-end model, based on deformable transformers~\citep{Vaswani2017,Deformable}, simplifying the process while ensuring the online generation of accurate and coherent captions.
Specifically, our approach is based on the deformable transformer of \citet{Deformable}, which is a model that was introduced as an architecture to improve the performance of object detectors by attending to sparse spatial locations and incorporating multi-scale feature representations.
We have adapted this deformable transformer architecture to process video sequences for the novel problem of LVC.
Our localization and captioning modules process the streaming video, producing online dense captions that are enhanced with an extra filtering process.

Similar transformers-based architectures were also used in offline dense video captioning models, e.g. \citep{Yang2023, wang2021}.
For instance, while both our model and~\citep{wang2021} leverage deformable transformers, their architectural designs reflect fundamentally different objectives due to the contrasting demands of their respective tasks.
In general, the main differences with respect the previous offline dense video captioning models \citep{Yang2023, wang2021}, that also use transformer-based architectures are as follows.
First, they assume full access to the entire video sequence during both training and inference.
This enables the models to perform holistic event localization and captioning by leveraging global temporal information.
In contrast, our LVC model processes partial, sequentially available video segments under causal constraints, where no future frames are accessible.
To accommodate this, our deformable transformer is modified to focus solely on the features within the current segment and refine predictions dynamically as new frames arrive.
Second, in the offline dense video captioning setting, a multi-scale deformable attention can be used to process densely extracted frame features from the entire video.
This attention mechanism ensures that the model captures relationships between temporally distant events.
In our LVC model, we adapt the multi-scale deformable attention mechanism to operate on temporally truncated input streams.
For instance, our event queries are tailored for online processing, designed to anticipate and refine event boundaries dynamically as frames are processed.
This iterative refinement aligns with the causal nature of the LVC task, allowing the model to balance precision with latency.
We also add a temporal filtering mechanism, integrated with the learnable event queries, to dynamically generate the dense captions for the events within the limited temporal context available, enabling precise and actionable online predictions.
These distinctions highlight how our use of deformable transformers is uniquely adapted to the challenges of LVC.

In table \ref{table:review_models_dvc} we provide a global comparison of the main models for dense video captioning. Note that only our approach has been specifically designed for the novel LVC problem.

\begin{table}[t]
\centering
\resizebox{\linewidth}{!}{
\begin{tabular}{l|cccr}
\toprule
\textbf{Work} & \textbf{Joint Loc. \& Cap.} & \textbf{LVC} & \textbf{Contributions} \\
\midrule
\citep{krishna2017dense} & \xmark & \xmark & First work; Proposals module + attention-based LSTM for caption generation. \\
\citep{wang2018bidirectional,yang2018hierarchical} & \xmark & \xmark & Enriched event representations by context modeling.\\
\citep{wang2020event} & \xmark & \xmark & Incorporate event-level relationship.\\
\citep{Iashin2020MDVC,Yang2023}  & \xmark & \xmark & Use of multi-modal feature fusion.\\
\midrule
\citep{li2018jointly} & \checkmark & \xmark & Use of proxy task based on language rewards to enhance the localization.\\
\citep{wang2021} & \checkmark & \xmark & Inter-task interactions with deformable transformers.\\
\citep{zhou2018end} & \checkmark & \xmark & Differential masking for linking the gradient flow from captioning loss to proposals' boundaries.\\
\midrule
LVC (Ours) & \checkmark & \checkmark & Deformable transformers + temporal filtering for video streams.\\ \bottomrule
\end{tabular}
}
\caption{Review of other models for the problem of dense video captioning. We organize the literature based on whether the models address the problems of localization and captioning jointly, and whether they can handle the proposed LVC scenario or operate offline. We also summarize some of the main contributions of the works.}
\label{table:review_models_dvc}
\end{table}

\section{Live Video Captioning}
\label{sec:approach}

\subsection{Live Video Captioning: problem formulation}\label{sec:lvc_formulation}
We define the problem of Live Video Captioning (LVC) as the process of obtaining dense captions for a video stream as soon as the video frames are available.
Unlike the traditional video captioning scenario, which we refer to as offline video captioning, we do not have access to the entire video for analysis.
Instead, LVC models can only access the content coming from a video stream up to the time instant $t$ to generate caption predictions for that instant.
In other words, LVC models process the video in an online fashion, implying that the dense caption generation system has access to the current information of the video and past information, but never future information.
Therefore, these systems are inherently causal.

LVC models face unique challenges when compared to traditional offline captioning systems. In offline systems, the model has access to the entire video, allowing for accurate temporal segmentation and caption generation after full observation.
In contrast, LVC models must generate captions in an online fashion, working with partial, often incomplete frames, and make predictions with limited information. This online constraint introduces several challenges: (i) accurate temporal segmentation with minimal context, (ii) the need for anticipation of actions before they are fully visible, and (iii) ensuring that caption generation remains both fast and precise.

We must assume that for LVC solutions, there is always information yet to be revealed.
Caption predictions are made based on partial content of the video, in case it is already recorded.
However, it is in the context of live video streams where LVC models gain special relevance.
We can think of the following applications and problems.
For instance, consider a surveillance camera monitoring a busy public space, where the flow of events is unpredictable. In such a scenario, we can never be certain about what might occur next—a sudden altercation, a person leaving an unattended bag, or an individual entering a restricted area. However, it is crucial to generate dense captions as soon as the visual information is available to the system. This capability enables timely detection and response to critical events, enhancing safety and situational awareness. Another illustrative example is a robot equipped with a camera, tasked with generating detailed descriptions of the scenes it encounters while navigating a dynamic environment, such as a bustling factory floor or a crowded shopping mall. The robot might describe activities like workers assembling parts, customers interacting with products, or obstacles in its path. This continuous scene interpretation is vital for tasks like human-robot interaction, navigation, and situational understanding. Any traditional offline model for dense video captioning would not be able to generate these descriptions, as these models have been trained to operate only when the entire video is available. For instance, models that rely on action proposals (e.~g. \citep{krishna2017dense,wang2018bidirectional,yang2018hierarchical}) would be severely affected. In the novel LVC paradigm, it is not desirable to wait until the action is finished to have a correct caption prediction for it.

Therefore, a critical advantage of Live Video Captioning (LVC) is its ability to anticipate actions and provide temporally precise localization of events, a limitation commonly observed in offline dense video captioning methods. For instance, in a video where a person begins to raise a glass to drink, followed by a second person setting a plate on the table, offline methods would generate overlapping captions like '[2.0s-6.0s] A person moves' and '[5.0s-9.0s] A person interacts with an object,' failing to distinguish between the two actions clearly or provide precise temporal boundaries. By contrast, LVC models would produce '[2.0s-4.0s] A person raises a glass' and '[4.0s-6.0s] Another person places a plate on the table,' with non-overlapping, accurate temporal segmentation. This novel scenario not only allows for early recognition of ongoing events but also improves temporal granularity, enabling a clearer distinction between simultaneous or sequential actions. These capabilities are vital for applications requiring immediate and accurate action detection, such as the monitoring systems or interactive robotics described.

In the new context of LVC, we proceed to define the fundamental properties that characterize it, which are as follows:

\begin{enumerate}
 \item \textbf{Input assumption}: Streaming videos are assumed to be the natural inputs for LVC approaches, where neither length nor content of the \emph{entire} video are accesible.
 \item \textbf{Timeliness}: Captions must be generated as soon as the actions unfold.
 \item \textbf{Causality}: Dense caption generation must be causal, so future frames cannot be used.
 \item \textbf{Temporal adjustment}: The caption prediction must be adjusted to the temporal information available up to the time corresponding to the prediction instant.
 \item \textbf{Irreversibility}: No post-processing or subsequent thresholding of caption scores can be applied once they are generated for a previous instant of time. LVC methods cannot revise past generated captions.
\end{enumerate}

\subsection{Our Live Video Captioning Model}\label{sec:lvcmodel}

For the implementation of our LVC model, we drew inspiration from the latest advances in solutions for offline dense video captioning~\citep{Yang2023, wang2021}, where transformer-based architectures~\citep{Vaswani2017} were used.

\begin{figure}[ht!]
    \centering
    \includegraphics[width=0.5\linewidth]{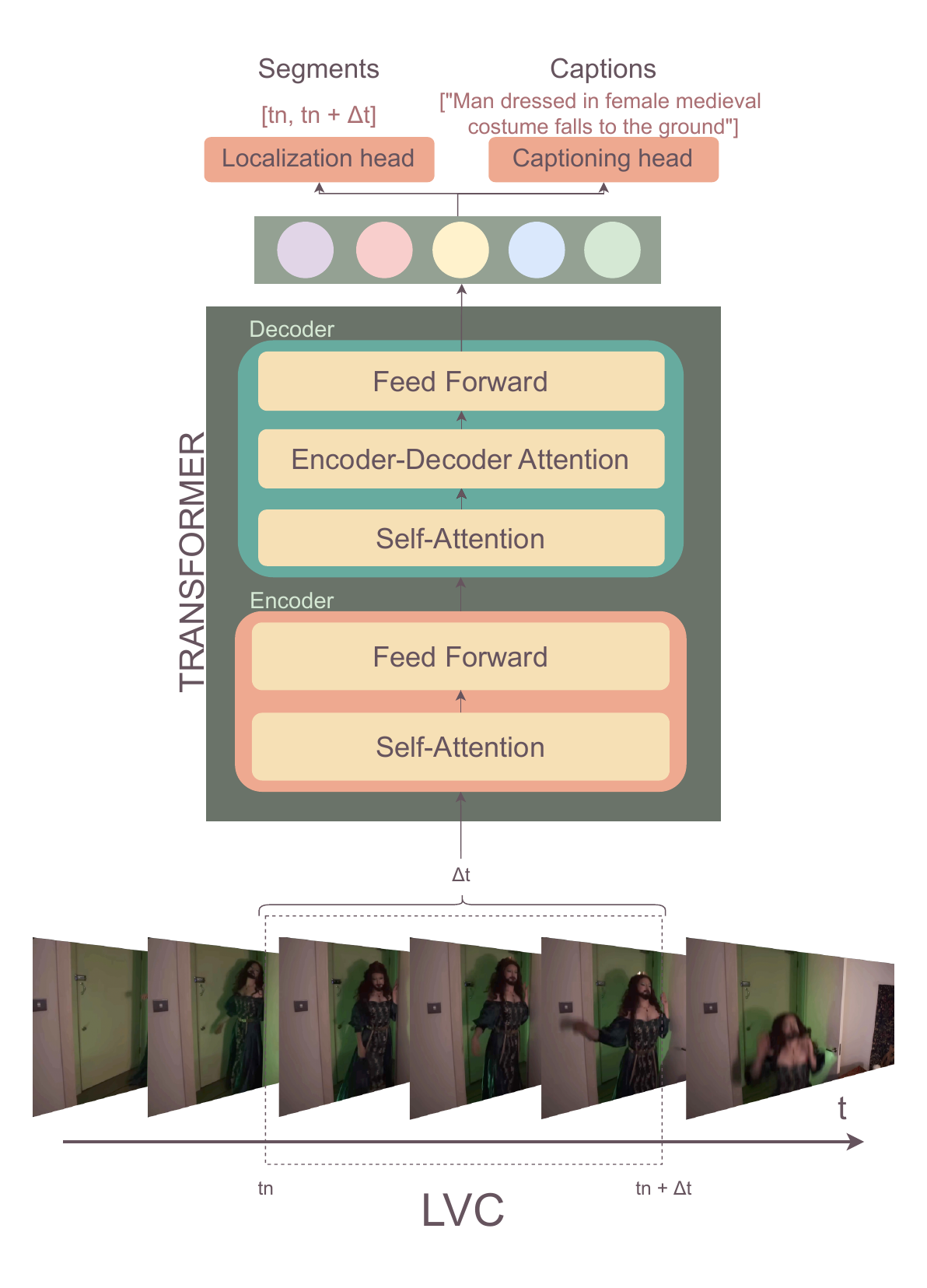}
    \caption{LVC adopts a deformable \textit{transformer}-based architecture to learn the interaction of different frames of the video, including learnable event queries to capture the significance of the relationship between frames and events. Two prediction heads run in parallel on the query features, leveraging mutual benefits between the two tasks and improving their performance together.}
    \label{fig:lvc_model}
\end{figure}

As it is shown in Figure~\ref{fig:lvc_model}, we develop a deformable transformer model applied to the novel online dense video captioning problem.
The deformable transformer model was introduced in~\citep{Deformable} as an architecture to improve the performance of
object detectors by attending to sparse spatial locations and incorporating multi-scale feature representations.
Technically, our LVC model integrates the deformable transformer-based architecture to construct online caption predictions, taking temporal video segments of length $\Delta t$ as inputs.
 Given an input video stream $V_i = \left\{I_1, I_2, I_3, \ldots \right\}$, we first split it in video segments $S_i$
of duration $\Delta t$, hence $V_i = \left\{S_1, S_2, S_3, \ldots \right\}$.
Note that our LVC does not have to access the entire video, as required for offline systems such as~\citep{
    show_tell_sum, efficient, Yang2023, wang2021}.

These video snippets $S_i$ are the inputs for a deformable transformer model, with the corresponding encoder and decoder.
The following operational scheme is followed from the introduction of the video segment to the model until the captions are obtained.
First, our model extracts the features for each of the frames in the video segment $S_i$.

For our goal of generating dense captions for the actions and events occurring in videos, we have chosen to use a pre-trained model for action recognition.
In particular, for our experiments, we used the Temporally-Sensitive Pretraining (TSP) feature extractor~\citep{tsp}.
However, our LVC model can be integrated with any action recognition feature extractor.
We then rescale the temporal dimension of the feature maps obtained to a fixed length $T$ using interpolation.
To effectively use multi-scale features for forecasting events at different scales, we incorporate $L$ temporal convolutional layers with a stride of 2 and a kernel size of 3.
This approach generates feature sequences at various resolutions, ranging from $T$ to $\frac{T}{2^L}$.
These multi-scale features $\{\mathbf{x}^l\}_{l=1}^L$, combined with their positional embeddings, are input into the deformable transformer encoder, which captures frame-to-frame relationships across multiple scales.

Note that, accurate captioning for videos with multiple simultaneous events relies heavily on temporal information to segment overlapping actions and on contextual information to maintain coherence.
Our model incorporates these elements through the proposed deformable transformer-based architecture.
Temporal information is leveraged via multi-scale temporal convolutional layers that enable the extraction of features at varying time resolutions, capturing the precise start and end of events. Contextual relationships are modeled using the deformable attention mechanism, which dynamically attends to relevant temporal regions, allowing the system to refine predictions based on localized action patterns.
These components work in tandem to disambiguate concurrent events and generate distinct, temporally aligned captions, even in complex multi-event scenarios.
Moreover, our model integrates the multi-scale temporal convolutional layers to aggregate features across consecutive frames, enabling the detection of temporal patterns that extend beyond individual frame boundaries.

Our deformable transformer, as described in~\citep{Deformable}, is an encoder-decoder framework that utilizes multi-scale deformable attention (MSDeformAttn).
For our set of multi-scale feature maps $\{\mathbf{x}^l\}_{l=1}^{L}$ where $\mathbf{x}^l \in \mathbb{R}^{C\times H_l\times W_l}$, a query element $\mathbf{q}_j$, and a normalized reference point $\mathbf{p}_q\in[0, 1]^2$, MSDeformAttn produces a context vector via a weighted sum of $K \times L$ sampling points across the feature maps at $L$ scales:
\[
  \begin{aligned}
    &\text{MSDeformAttn} (\mathbf{q}_j, \mathbf{p}_j, \{\mathbf{x}^l\}_{l=1}^{L}) = \sum_{l=1}^L \sum_{k=1}^K A_{jlk} \mathbf{W} \mathbf{x}^l_{\Tilde{\mathbf{p}}_{jlk}} \; ,\\
    &\Tilde{\mathbf{p}}_{jlk} = \phi_l(\mathbf{p}_j) + \Delta \mathbf{p}_{jkl}.
  \end{aligned}
\]

In this equation, $\Tilde{\mathbf{p}}_{jkl}$ and $A_{jkl}$ denote the position and attention weight of the $k$-th sampled key at the $l$-th scale for the $j$-th query element, respectively. $\mathbf{W}$ represents the projection matrix for key elements, and $\phi_l$ maps the normalized reference points into the feature map at the $l$-th level.
The sampling offsets $\Delta \mathbf{p}_{jkl}$ are relative to $\phi_l(\mathbf{p}_j)$.
Both $A_{jkl}$ and $\Delta \mathbf{p}_{jkl}$ are determined through linear projection onto the query element.
Overall, in our deformable transformer, self-attention modules in the transformer encoder and cross-attention modules in the transformer decoder are replaced with deformable attention modules.

The decoding network comprises a deformable transformer decoder and three parallel components, leveraging the
strategy in~\citep{wang2021}: a captioning head for generating captions, a localization head for predicting event boundaries with confidence scores, and an event counter for estimating the number of events.
The decoder's objective is to directly query event-level features from the frame features using $N$ learnable
embeddings (referred to as event queries) $\{\mathbf{q}_j\}_{j=1}^N$ and their associated scalar reference points ${p}_j$. The reference point ${p}_j$ is obtained through a linear projection followed by a sigmoid activation applied to ${\mathbf{q}}_j$. These event queries and reference points act as initial estimates for the events' features and locations (center points) and are iteratively refined at each decoding layer. The refined query features and reference points are denoted as ${\tilde{{\mathbf{q}}}_j}$ and ${\tilde{{\mathbf{p}}}_j}$, respectively.

Our \textbf{localization head} produces a box prediction and a binary classification for each event query.
The box prediction task aims to determine the 2D relative offsets (center and length) of the ground-truth segment with respect to the reference point.
Binary classification generates the foreground confidence for each event query. Both the box prediction and binary classification are carried out using multi-layer perceptrons. This process results in a set of tuples
$[t_{j1}, t_{jf}, \alpha^{\rm loc}_j]_{j=1}^{N}$ that represent the detected events, with $t_{j1}$ and $t_{jf}$ the
initial and final times, and where $\alpha^{\rm loc}_j$ is
the localization confidence of the event query ${\tilde{{\mathbf{q}}}_j}$.

Instead of using a two-stage scheme, our LVC employs enhanced event query representations in parallel localization and
captioning heads, allowing these two subtasks to be closely related.
LVC directly produces a set of events with an appropriate size without relying on heuristic techniques to eliminate redundancy.
Within our deformable transformer (see Figure \ref{fig:lvc_model}), \textit{Event Queries} are produced and
introduced into the decoder.
Each of these queries will result in a prediction for a caption.
For all experiments, we use a total of 10 queries, although in the ablation studies we evaluate the impact of this parameter.

Our \textbf{captioning head} feeds ${\tilde{{\mathbf{q}}}_j}$ into a vanilla LSTM at each timestamp.
The word $w_{jt}$ is predicted by a fully connected layer followed by a softmax activation over the hidden state $\mathbf{h}_{jt}$ of the LSTM.
Note that this LSTM operates only within the video segment of duration $\Delta t$ that our LVC model is analyzing.

The \textbf{event counter head} predicts the number of events using a max-pooling layer and a fully connected layer with softmax activation, producing a fixed-size vector \(\mathbf{r}_{\rm len}\), where each value refers to the possibility of a specific number. The predicted event number is obtained by \(\text{argmax}(\mathbf{r}_{\rm len})\). Top \(N_{\rm set}\) events are selected based on accurate boundaries and captions.

Confidence for each event query is calculated by combining two components: location confidence and a modulated caption confidence.
The location confidence evaluates the spatial accuracy of the event, while the modulated caption confidence incorporates adjustments for variability in sentence length.
During training, our LVC model generates a set of \(N\) predicted events, each comprising a spatial location and an associated caption.
To associate these predictions with the ground truth, the Hungarian algorithm is employed, solving a bipartite matching problem that optimizes a comprehensive cost function.
This cost function integrates two primary components: the Generalized Intersection over Union (GIoU) loss~\citep{giou} for spatial alignment and a focal loss~\citep{focalloss} for classification accuracy.

Our total training loss, \(\mathcal{L}_{\text{total}}\), is computed as a weighted combination of several loss terms:
$\mathcal{L}_{\text{total}} =  \lambda_{\text{GIoU}} \mathcal{L}_{\text{GIoU}} + \lambda_{\text{cls}} \mathcal{L}_{\text{cls}} + \lambda_{\text{count}} \mathcal{L}_{\text{count}} + \lambda_{\text{cap}} \mathcal{L}_{\text{cap}}$, where \(\mathcal{L}_{\text{GIoU}}\), \(\mathcal{L}_{\text{cls}}\), \(\mathcal{L}_{\text{count}}\), and \(\mathcal{L}_{\text{cap}}\) represent the GIoU loss, classification loss (i.e. focal loss), counting loss, and caption loss, respectively.
Our counting loss is the cross-entropy loss between the estimated count distribution and the one in the ground-truth.
For the captioning loss we use the cross-entropy computed between the word probability estimated by our model and the ground-truth normalized by the length of the caption.
The hyperparameters \(\lambda_{\text{GIoU}}\), \(\lambda_{\text{cls}}\), \(\lambda_{\text{count}}\), and \(\lambda_{\text{cap}}\) control the relative contributions of each loss term.
For our experiments we use \(\lambda_{\text{GIoU}} = 2\), and \(\lambda_{\text{cls}} = \lambda_{\text{count}} =\lambda_{\text{cap}} = 1\).

While offline models for dense video captioning can access all available information by analyzing the entire video
before generating the captions, our LVC model only has access to the information available in the video segment of duration $\Delta t$.
This limitation can hinder the quality of the subtitles because a short video segment might not provide enough context.
In our LVC model, captions are generated in an online manner based exclusively on the video data available up to the current time step.
Once a caption is produced for a given video segment, it remains immutable and is not revised as new frames are processed.
This design choice enforces the causality constraints essential to the LVC task, ensuring that each caption accurately reflects the observed content at the moment of generation without relying on future information.
This approach distinctly differentiates LVC from traditional offline methods, which have access to the entire video sequence and can even refine captions retrospectively.

\begin{figure}[ht!]
    \centering
    \includegraphics[width=\linewidth]{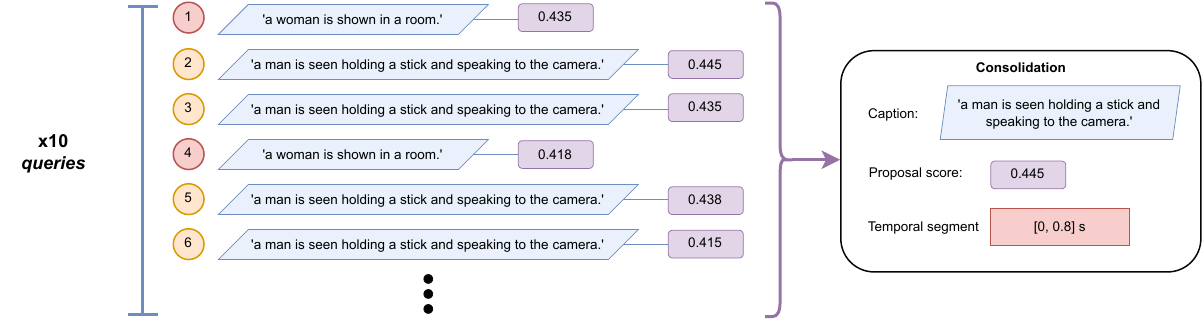}
    \caption{Example of caption consolidation for a video segment. }
    \label{fig:Caption_consolidation}
\end{figure}

Therefore, to improve the quality of the captions, we have included a \textbf{temporal filtering for the caption consolidation}, as shown in Figure~\ref{fig:Caption_consolidation}.
First, our deformable transformer uses a query mechanism comprising $N$ queries per video segment of duration $\Delta t$.
Subsequently, our filtering module accepts the predictions generated by these event queries, $\{\mathbf{q}_j\}_{j=1}^{N}$, as input.
Each query produces a predicted caption $c_j$, an associated temporal segment $[t_{j1}, t_{jf}]$, and a confidence score $\alpha_j$, resulting in a set of tuples $\{(t_{j1}, t_{jf}, c_j, \alpha_j)\}_{j=1}^N$.
Considering the temporal overlap between the predictions, the consolidation module orchestrates a voting mechanism to determine the final captions for the given video segment.
Specifically, the temporal filtering we propose consists in a module that evaluates the frequency of occurrence of each overlapped predicted caption  among the set of predictions. Let $f(c)$ represent this frequency for a particular caption $c$, and $w(c)$ its maximum score.
The ultimate caption $c_*$ for the set of temporally overlapped predictions  is selected as the one with the highest frequency: $c_* = \arg\max_{c \in \{c_1, \dots, c_N\}} f(c)$. Being the final tuple for the prediction as follows: $\{(t_{*1}, t_{*f}, c_*, w(c_*))\}$.

\subsection{Novel Evaluation Metric for Live Video Captioning: the Live Score}
\label{sec:novel_metric}

Dense video captioning models, by operating in offline mode, have traditionally been evaluated using offline metrics.
The new LVC paradigm we propose necessitates the development of a new evaluation metric with an online nature.
We begin by describing the typical offline evaluation scheme and then highlight its main limitations for LVC systems, thereby introducing the properties that an online evaluation metric must possess.
Next, we introduce the formulation associated with the new proposed metric, the Live Score, including all of its variants.

\subsubsection{Online evaluation metric properties}\label{sec:metric_online_properties}

The main characteristics of traditional offline dense video captioning metrics are as follows.
All the information from the video annotations is introduced at once in the metrics.
Then, the scorer rates the entire input video regardless of its duration.
This scenario is quite different from that of LVC, where dense caption predictions with their associated timestamps arrive in temporal blocks, whose duration is less than that of the full video being processed.
Moreover, the offline metrics cast a score that corresponds to the average score for a full video.
This fact does not allow us to \emph{observe} the temporal evolution in the performance of the models, and this property is interesting for LVC.
In other words, we need an online nature metric that evolves with the video and reflects the online accuracy of LVC systems.

Therefore, the characteristics that an online metric for LVC must fulfill are as follows:
\begin{itemize}
    \item \textbf{Video stream-based scoring}: Ability to obtain scores from a video stream, therefore no access to a whole video is needed.
    \item \textbf{Causality compliance}: The metric should not have access to future information, only to what is being processed at the moment and what has already been processed.
    \item \textbf{Memory-aware}: Implementation of a record of the scores obtained in the video. This will be used to calculate subsequent results and allow for an evolution over time.
\end{itemize}

\subsubsection{The Live Score}\label{sec:tls}

For the new paradigm of LVC, we propose a purely online metric: the Live Score (LS).
In short, it is an adaptation of the various scores used in offline metrics, but tailored to process video streams online, causally, and considering the history of the video, as we have specified in the previous section.
Evaluating captioning quality in live video scenarios requires continuous assessment, as the accuracy of captions must be ensured at each stage of video playback.
We propose an experimental evaluation guided this novel LS metric, which tracks the temporal evolution of caption accuracy as the video progresses. The LS metric allows us to evaluate caption performance at each moment, taking into account both the immediate accuracy and the cumulative quality of predictions. This continuous evaluation is crucial in live scenarios, where the model must maintain high performance throughout the video’s duration, ensuring that captions remain relevant and precise as events unfold.

We begin with the necessary mathematical formulation to define the LS.
Let an LVC model aim to produce a series of caption predictions by analyzing an input video stream every $\Delta t$ seconds.
Note that $\Delta t$ will be the only configurable parameter of the new metric LS.
We define $C_i$ as the set of captions generated by the model LVC when presented with a video $V_i$:
\begin{multline}\label{eq:LVC}
    LVC(\Delta t, V_i) \Rightarrow C_i = \{[t_{1i},t_{1f},c_1,\alpha_1],[t_{2i},t_{2f},c_2,\alpha_2], \dots, \\
    [t_{ni},t_{nf},c_n,\alpha_n]\}\; ,
\end{multline}
where $t_{ni}$ and $t_{nf}$ are the start and end times of each timestamp, respectively, $c_n$ contains the predictions for the captions, and $\alpha_n$ encodes the confidence assigned by the LVC model to each caption.

\begin{figure}[t]
    \centering
    \includegraphics[width=0.8\linewidth]{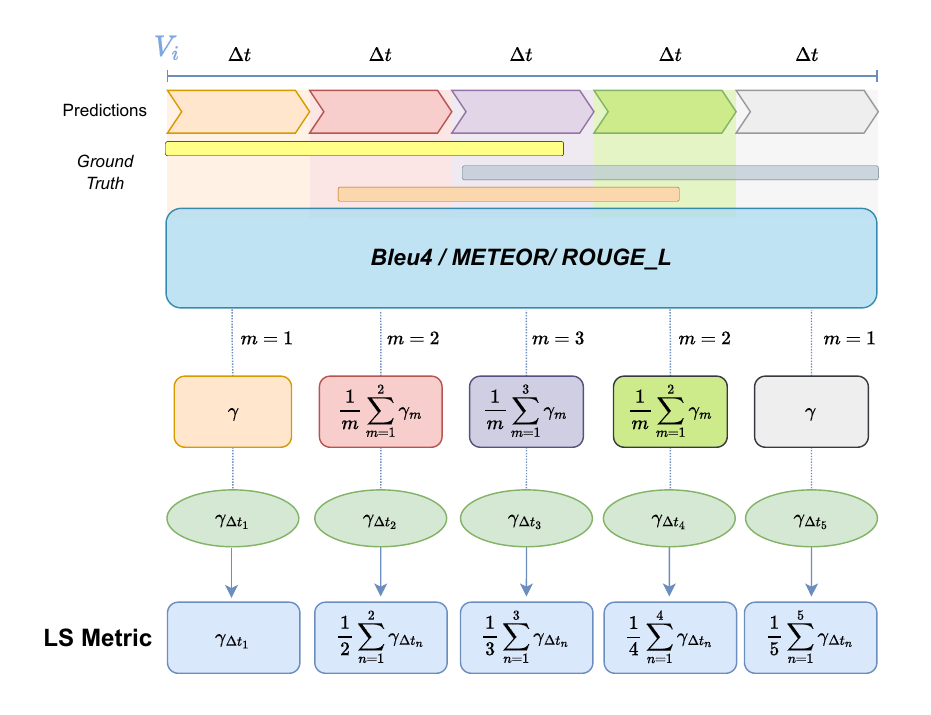}
    \caption{The LS metric. It allows for an online and continuous evaluation of a video stream, analyzed every $\Delta t$ seconds. Our metric allows for the integration of any scorer (e.g. METEOR, Bleu4 or Rouge-L) in the online or live evaluation.}
    \label{fig:met_online}
\end{figure}

The LS metric will process the data in $C_i$ online, providing a score $\gamma_{t'}$ for each timestamp $t'$, with a resolution of $\Delta t$ seconds.
We propose to combine our LS metric with any of the traditional scorers for video captioning (see Figure~\ref{fig:met_online}). This scorer is now evaluated continuously, and our LS metric allows for the observation of its evolution, instantaneously.
The scorers are responsible for comparing the similarity between the predicted and annotated captions.
The ones we have integrated into our LS metric and that have been used in experiments are as follows:
\begin{itemize}
    \item METEOR \citep{denkowski:lavie:meteor-wmt:2014}: It is an automatic translation quality evaluation metric that calculates word, phrase, and synonym similarity scores.
    \item Bleu4 \citep{BLEU}: Automatic evaluation metric used in text generation and machine translation. It emphasizes the accuracy of matching four-word \textit{n-grams} and evaluates the similarity between the generated output and human references using the count of matching \textit{n-grams}.
    \item Rouge-L \citep{Lin2004ROUGEAP}: Automatic evaluation metric primarily used in automatic text summarization. The generated summary and the reference summary are compared using word count and summary length.
\end{itemize}

To reflect a continuous temporal evolution, when we have the score calculated at $t'$, we compute the mean with all previous scores, so that the LS metric is formulated as follows:
\begin{equation}
\text{LS}(t',LVC(\Delta t, V_i)) = \frac{\displaystyle\sum_{n = 1}^{K}\gamma_{t'_n}}{K} \; ,
\end{equation}
where $t_n^{'} = n \cdot \Delta t$, the numerator corresponds to the sum of all scores calculated up to the current moment $t'$, and $K = \frac{t'}{\Delta t}$.

As it is shown in Figure~\ref{fig:met_online}, we can have multiple ground-truth captions associated with the video segment we are processing.
Remember we are dealing with the \emph{dense} video captioning problem, hence this situation is possible.
Our metric will produce a score between the predictions and each annotated caption, resulting the final score $\gamma_{t'}$ as the average of all generated scores for that segment.
We show in Figure~\ref{fig:met_online} a graphical example, where the LS metric is used to process a video $V_i$ segmented into fragments with length $\Delta t$, each one containing $m$ associated captions.

The proposed LS metric has an online nature, meeting all the requirements detailed in Section~\ref{sec:metric_online_properties}.
However, it does not take into account the influence of false positives, i.e., predictions of captions that do not appear in the annotations available in the database.
In other words, the proposed metric does not include any calibration mechanism with respect to false positives.
In a realistic scenario for the LVC problem, such as generating captions for a live video stream from a surveillance camera, it is highly likely that there are large portions of the video where no action is occurring.
Thus, in an LVC model applied to a video surveillance system, we must avoid at all costs the model generating captions for events that have not occurred.
Indeed, false-positive captions pose a significant challenge in live video captioning where they can disrupt the user experience by generating frequent, inaccurate descriptions.
An LVC system will be accurate if it provides accurate captions, but also if it only provides captions when something relevant is happening in the video.

To calibrate our metric and make it sensitive to false positives, we propose integrating a penalty for false positives into the LS.
This new version of the metric is called weighted-LS (wLS), and is formulated as follows:
\begin{equation}
\text{wLS}(t',LVC(\Delta t, V_i)) = \frac{\displaystyle\sum_{n = 1}^{K}\gamma_{t'_n}}{K} \cdot e^{-\beta}\; ,
\end{equation}
\begin{equation}
\beta = \frac{\displaystyle\sum_{n = 1}^{K}fp(t_n')}{K}\; ,
\end{equation}
where we have added the correction factor $\beta$, dependent on $fp(t_n')$, which is the number false positives corresponding to the video segment associated to $t_n'$.
By incorporating this weighted metric, our experimental evaluation pipeline enhances the reliability of live captioning systems, ensuring a more seamless and coherent user experience.

\begin{figure}[t]
    \centering
    \includegraphics[width=0.8\linewidth]{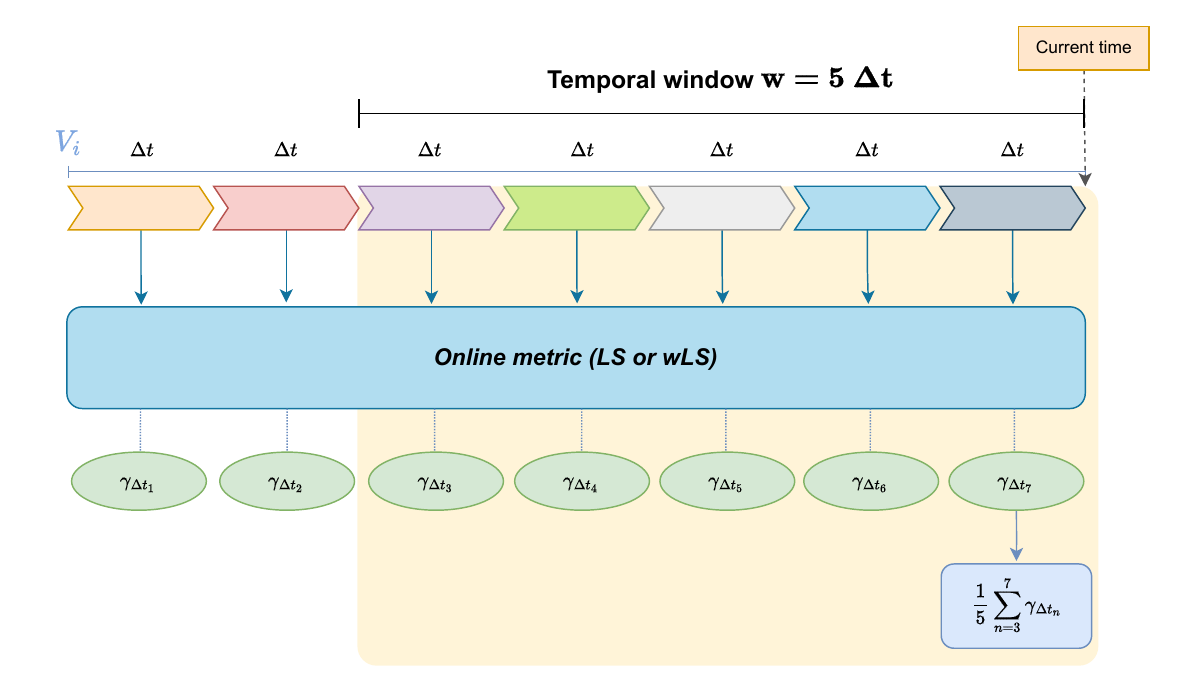}
    \caption{Operation of the online metric with fixed temporal window history. We observe how that temporal window moves along the video timeline. The window, with size $w = 5$, encompasses the scores that will be considered to compute the score associated with the current instant. The first two slots have been discarded. The diagram has been simplified for ease of understanding, but the calculation of scores for each $\Delta t$ is the same as in the previous scenarios.}
    \label{fig:metric_online_history}
\end{figure}

The two new proposed metrics, LS and its calibrated version wLS, allow for the online and continuous evaluation of what happens in a video up to time $t'$, considering the entire history of the video from $t=0$.
It may happen that the evaluation process starts with predictions that have very low scores and gradually improve over time, or vice versa.
In such scenarios, since the metric is calculated based on \emph{all} previous scores obtained by the system from $t=0$, it always considers the entire temporal timeline, and the metric may fail to reflect the system's most \emph{recent} behavior.
To address this issue, we propose a version for both LS and wLS that considers only a fixed temporal history window for computing the performance of the LVC system at time $t'$, covering only the interval $[t'-w\Delta t,t']$.
Here, $w$ defines the size of this fixed temporal window used to compute the metrics LS and wLS.
By updating the metric considering only the fixed temporal history window, we allow it to evolve, reflecting the current performance of the LVC model.
In Figure~\ref{fig:metric_online_history}, we illustrate the calculation process for a temporal window size of $w = 5$.

Incorporating the fixed temporal history window, the formulations of the previously described metrics are as follows:
\begin{equation}
    \text{hLS}(t',LVC(\Delta t, V_i)) = \frac{\displaystyle\sum_{n = \text{max}(1, K+1-w)}^{K} \gamma_{t'_n}}{K}\; ,
\end{equation}
\begin{equation}
    \text{hwLS}(t',LVC(\Delta t, V_i)) = \frac{\displaystyle\sum_{n = \text{max}(1, K+1-w)}^{K} \gamma_{t'_n}}{K} \cdot e^{-\beta}\; ,
\end{equation}
\begin{equation}
    \beta = \frac{{\displaystyle\sum_{n=\text{max}(1, K+1-w)}^{K}fp(t_n')}}{K} \;.
\end{equation}

\section{Experimental Evaluation}
\label{sec:experiments}

In this section, we provide details of the experimental evaluation designed for the proposed novel LVC problem.

We start in Section~\ref{sec:experimental_setup} with a description of the experimental setup, where we outline the database used and the adaptations made to it for the evaluation of the live models.
Subsequently, in Sections~\ref{sec:ev_orig}-\ref{sec:demo}, we include both qualitative and quantitative results of all our experiments.
The questions we want to address with the proposed experimental evaluation are as follows:
\begin{enumerate}
    \item Are offline experimental evaluation environments adequate for LVC models?
    \item Are the proposed new metrics suitable for the LVC problem, and do they allow to judge the temporal evolution of LVC systems?
    \item What is the performance of the proposed model for LVC? With respect to the state of the art in offline models, how does the proposed LVC approach perform?
\end{enumerate}

\subsection{Experimental setup}
\label{sec:experimental_setup}

We have used for our experimental evaluation the \textit{ActivityNet Captions}~\citep{krishna2017dense} dataset.
This database is actually a subset of data from \textit{ActivityNet}~\citep{caba2015activitynet}.
Specifically, \textit{ActivityNet Captions} consists of a set of 20,000 videos totaling 849 hours of video with a total of 100,000 descriptions, each with its start and end timestamp.
On average, each annotated video contains 3.65 localized phrases, with each phrase averaging 13.48 words.
The videos were generated at a rate of 30 frames per second.
For our experiments we have chosen the validation set, which contains 4,926 videos.

The different $\Delta t$ values used in the experiments were $(24, 48, 72, 96, 120, 150)$ in frame numbers.
We justify this choice of values for $\Delta t$ because all of them represent a reasonable temporal length for online applications.
If the values were greater, the delay between predictions would be too high, making them unsuitable for consideration as \emph{live} models.

\subsection{Evaluating LVC models with off-line metrics}\label{sec:ev_orig}

Are \emph{offline} experimental evaluation metrics suitable for the novel LVC problem?
This is the question we want to specifically address in this section.
We set up an experimental evaluation scenario in which we use the official \emph{offline} metrics provided in the ActivityNet Challenge 2018~\citep{activitynet2018summary}, on the online predictions generated by our LVC system.
For localization performance, the average precision and average recall across intersection over union at different thresholds are used.
For dense captioning performance, the official evaluation tool provided by ActivityNet Challenge 2018 is followed, which calculates the average precision measured by BLEU4, METEOR, and ROUGE\_L scorers, of the matched pairs between generated captions and the ground truth across intersection over union thresholds of 0.3, 0.5, 0.7 and 0.9.

In Table~\ref{table:evaluation_gt_orig}, we first present the results in terms of dense captioning accuracy.
We provide a detailed comparison between the results of our online LVC model and those offered by the offline models PDVC~\citep{wang2021} and Vid2Seq~\citep{Yang2023}, that are the state of the art for the dense video captioning problem.

\begin{table}[ht!]
\centering
\resizebox{\linewidth}{!}{%
\begin{tabular}{@{}lccc|ccc@{}}
\toprule
\textbf{Model} & \textbf{Live} & $\mathbf{\Delta t}$ & \textbf{Features} & \textbf{Bleu4} & \textbf{METEOR} &  \textbf{ROUGE\_L} \\
\midrule
\multirow{3}{*}{PDVC~\citep{wang2021}} & \multirow{3}{*}{\xmark}
              & & C3D      & 1.65  & 7.50   &  - \\
              & & & TSN      & 1.78  & 7.96   &  - \\
              & & & TSP      & \textbf{2.17}  & \textbf{9.03} &  -\\
\multirow{3}{*}{PDVC\_light~\citep{wang2021}} & \multirow{3}{*}{\xmark}
              & & C3D      & 1.51  & 7.11   &  - \\
              & & & TSN      & 1.66  & 7.97   &  - \\
              & & & TSP      & \textbf{1.77}  & \textbf{8.55}   &  - \\
Vid2Seq~\citep{Yang2023} & \xmark & & CLIP & -  & 8.5   &  -\\
\midrule
\multirow{6}{*}{LVC (Ours)} & \multirow{6}{*}{\checkmark}
 & 24  & \multirow{6}{*}{TSP}
                &  0.13 & 0.14   &     0.15 \\
 & & 48  &        &  0.47 & 0.45   &   0.51  \\
 & & 72  &        &  0.85 & 0.75   &   0.91 \\
 & & 96  &        &  1.21 & 1.03   &   1.29 \\
 & & 120 &        &  1.55 & 1.27   &   1.63  \\
 & & 150 &        &  \textbf{2.01} & \textbf{1.56}   &  \textbf{2.08}\\ \bottomrule
\end{tabular}
}
\caption{Experimental comparison in terms of accuracy of the captions generated by offline models and our online model LVC using traditional offline metrics for the \textit{ActivityNet Captions} dataset, using the validation set of videos.}
\label{table:evaluation_gt_orig}
\end{table}

Analyzing these results, we can draw the following conclusions.
The first one is that offline metrics favor offline models.
Our live system achieves low performance in some metrics.
The reason is clear: these metrics filter the captions generated by the models based on intersection-over-union thresholds, as we have seen, and the predictions of our LVC model are too short in temporal duration ($\Delta t$ is the value), so many of them do not survive this filtering and are naturally discarded by the offline metrics.
The second conclusion, related to the first one, is that offline metrics tend to improve as we increase the parameter $\Delta t$ in our LVC model.
The metrics were designed to work in offline scenarios where models can and should see the entire video first, and then generate all caption predictions.
This favors the generation of captions of much longer duration than those that can be generated by our LVC model (with a $\Delta t$ limit).
In fact, the offline dense captions can even occupy large temporal portions of the video.

We can also compare offline and live models in terms of the accuracy of temporal localization of the captions, again using traditional offline metrics: precision and recall.
In Table~\ref{table:scenario_offline}, we present this detailed analysis, and we can observe that the results of event localization obtained for our predictions using these offline metrics are not satisfactory.
Again, the metrics improve as $\Delta t$ increases in our LVC model.
The explanation is similar to what we have provided for the previous metrics: our caption predictions are associated with video segments of duration $\Delta t$, which causes them not to meet the intersection-over-union criteria employed by the offline metrics.
Observe the low performance when the threshold of 0.9 is used for the intersection-over-union.
\begin{table}[t]
\centering
\resizebox{\linewidth}{!}{%
\begin{tabular}{@{}lcc|ccccc|ccccr@{}}
\toprule
\multirow{ 2}{*}{\textbf{Model}} & \multirow{ 2}{*}{\textbf{Live}} & & \multicolumn{5}{c}{\textbf{Recall}} & \multicolumn{5}{c}{\textbf{Precision}}                    \\ \cmidrule(lr){4-8}
\cmidrule(lr){9-13}
 & & & 0.3   & 0.5   & 0.7   & 0.9   & avg   & 0.3   & 0.5   & 0.7   & 0.9   & avg
                    \\ \midrule
MFT~\citep{mft} & \xmark &   & 46.18 &  29.76 &  15.54 &  5.77  &  24.31 &  86.34 &  68.79 &
38.30 &  12.19 &  51.41 \\
SDVC~\citep{sdvc}& \xmark & & \textbf{93.41} &  \textbf{76.40} &  42.40 &  10.10 &  55.58 &  96.71 &  77.73 &
\textbf{44.84}&  10.99 &  57.57 \\
PDVC\_light~\citep{wang2021}& \xmark &    & 88.78 & 71.74 & \textbf{45.70} & \textbf{17.45} & \textbf{55.92} & 96.83 & 78.01 & 41.05 & \textbf{14.69} & 57.65 \\
PDVC~\citep{wang2021}& \xmark &           & 89.47 & 71.91 & 44.63 & 15.67 & 55.42 & \textbf{97.16} & \textbf{78.09} & 42.68
& 14.40 & \textbf{58.07} \\
Vid2Seq~\citep{Yang2023} & \xmark &  & - & - & - & - & 52.7 & - & - & - & - & 53.9 \\
\midrule
  & & $\mathbf{\Delta t}$ & & \\
\multirow{6}{*}{LVC - (ours)} & \multirow{6}{*}{\checkmark} & 24   & 7.60  & 2.72  &  0.89 &  0.16 & 2.84 & 1.81  &  0.45 & 0.21  &
 0.06 & 0.63 \\
& & 48   & 17.57 & 7.61  & 2.55  & 0.47  & 7.05 & 5.88  & 1.98  & 0.78  &
0.15  & 2.20                      \\
& & 72   & 25.40 & 12.30 & 4.61  & 0.92  & 10.81& 10.14 & 3.91  & 1.79  &
0.42  & 4.07                      \\
& & 96   & 31.76 & 16.45 & 6.42  & 1.31  & 13.98& 14.21 & 6.10  & 2.74  &
0.63  & 5.92                      \\
& & 120  & 36.70 & 19.34 & 7.67  & 1.74  & 16.36& 18.17 & 7.94  & 3.58  &
0.82  & 7.63                      \\
& & 150  & \textbf{42.50} & \textbf{22.87} & \textbf{9.67}  & \textbf{2.25}  & \textbf{19.32}& \textbf{23.11} & \textbf{10.24} & \textbf{4.77}  &
\textbf{1.23}  & \textbf{9.84}                      \\ \bottomrule
\end{tabular}
}
\caption{Caption localization for the validation video set of \textit{ActivityNet Captions}, using traditional offline metrics. Comparison with the state of the art offline models.}
\label{table:scenario_offline}
\end{table}

\subsection{Analysis using modified offline annotations}
One might argue that offline metrics could still be applicable in an online scenario if the video annotations are adjusted to align with the duration used by LVC models.
We have also performed this analysis for completeness, although we anticipate that this approach has significant drawbacks.

As an alternative to designing an online metric, one can attempt to use offline metrics but on a dataset where annotations have been modified to achieve an \emph{online appearance}.
In other words, it involves taking the temporal annotations for each caption and dividing them into small temporal segments that match the temporal window used by live video captioning models, i.e., $\Delta t$.

We have automated a process to modify the annotations provided in the validation set of the \textit{ActivityNet Captions} database.
Once the modified annotations are generated, traditional metrics for offline video captioning are employed, and the results are as follows.
In Table~\ref{tab:loc_captions_comparative}, we present the results in terms of caption localization in this new scenario and compare it with the performance obtained with the original annotations.
Note that by splitting the provided annotations, we ensure that the predictions of our live model are not filtered out because they do not meet the intersection over union criterion.

It is interesting to observe how adapting the annotations to the live solutions results in a considerable improvement in localization metrics.
The best average precision jumps from $9.84\%$ to $97.18\%$, while the best average recall reaches $90.16\%$ from only $19.32\%$.

\begin{table}[t]
\centering
\resizebox{\linewidth}{!}{%
\begin{tabular}{@{}c|ll|ccccc|ccccr@{}}
\toprule
&\multirow{2}{*}{\textbf{Model}} & \multirow{2}{*}{\textbf{$\mathbf{\Delta t}$}} &
\multicolumn{5}{c}{\textbf{Recall}} & \multicolumn{5}{c}{\textbf{Precision}}
               \\ \cmidrule(lr){4-8} \cmidrule(lr){9-13}
& & & 0.3   & 0.5   & 0.7   & 0.9   & avg   & 0.3   & 0.5  & 0.7   & 0.9   & avg
\\ \midrule
\multirow{6}{*}{\begin{tabular}[c]{@{}c@{}} Original \\ Annotations\end{tabular}}
& \multirow{6}{*}{LVC}
&   24    & 7.60  & 2.72  &  0.89 &  0.16 & 2.84 & 1.81  &  0.45 & 0.21  &  0.06
  & 0.63   \\
& & 48    & 17.57 & 7.61  & 2.55  & 0.47  & 7.05 & 5.88  & 1.98  & 0.78  & 0.15
& 2.20                      \\
& & 72    & 25.40 & 12.30 & 4.61  & 0.92  & 10.81& 10.14 & 3.91  & 1.79  & 0.42
& 4.07                      \\
& & 96    & 31.76 & 16.45 & 6.42  & 1.31  & 13.98& 14.21 & 6.10  & 2.74  & 0.63
& 5.92                      \\
& & 120   & 36.70 & 19.34 & 7.67  & 1.74  & 16.36& 18.17 & 7.94  & 3.58  & 0.82
& 7.63                      \\
& & 150   & 42.50 & 22.87 & 9.67 & 2.25 & \textbf{19.32} & 23.11 & 10.24 & 4.77 & 1.23 & \textbf{9.84}
\\ \midrule
\multirow{6}{*}{\begin{tabular}[c]{@{}c@{}}Modified \\ Annotations\end{tabular}}
& \multirow{6}{*}{LVC}
   & 24  & 90.97 & 90.08 & 89.24 & 88.32 & \textbf{90.16} & 97.63 & 97.32 & 96.85 & 96.23
& \textbf{97.18} \\
&  & 48  & 90.36 & 88.73 & 87.10 & 85.14 & 88.75 & 97.79 & 97.33 & 96.24 & 94.66
& 96.82 \\
&  & 72  & 89.85 & 87.46 & 85.05 & 82.30 & 87.46 & 97.88 & 97.22 & 95.46 & 93.18
& 96.39 \\
&  & 96  & 89.33 & 86.25 & 83.28 & 79.74 & 86.29 & 97.96 & 97.15 & 94.86 & 91.74
& 96.00 \\
&  & 120 & 88.88 & 85.02 & 81.48 & 77.43 & 85.16 & 97.99 & 97.01 & 94.06 & 90.42
& 95.57 \\
&  & 150 & 88.28 & 83.74 & 79.58 & 74.72 & 83.89 & 97.98 & 96.84 & 93.42 & 88.85
& 95.11
    \\ \bottomrule
\end{tabular}
}
\caption{Comparison of results using original annotations or annotations adapted to live models, in terms of caption localization.}
\label{tab:loc_captions_comparative}
\end{table}

We also report the performance of our model for the three different metrics used to evaluate the precision of the captions, see Table~\ref{tab:evaluation_gt_mod}.
Again, we compare the performance when the original annotations and the adapted ones are used.
We can observe that in this experiment the trend is that, the lower $\Delta t$ is, the higher the results are.

\begin{table}[t]
\centering
\begin{tabular}{@{}ll|ccc@{}}
\toprule
\textbf{Scenario}& $\mathbf{\Delta t}$& \textbf{Bleu4} &
\textbf{METEOR} & \textbf{ROUGE\_L} \\ \midrule
\multirow{6}{*}{\begin{tabular}[c]{@{}c@{}}Original \\ Annotations\end{tabular}}
 & 24   &  0.13 & 0.14   &   0.15  \\
 & 48   &  0.47 & 0.45   &   0.51  \\
 & 72   &  0.85 & 0.75   &   0.91  \\
 & 96   &  1.21 & 1.03   &   1.29  \\
 & 120  &  1.55 & 1.27   &   1.63  \\
 & 150  &  \textbf{2.01} & \textbf{1.56}   &  \textbf{2.08}  \\ \midrule
 \multirow{6}{*}{\begin{tabular}[c]{@{}c@{}} Modified \\ Annotations \end{tabular}}
 & 24   & \textbf{18.18} & \textbf{8.71}   &\textbf{18.27}    \\
 & 48   & 17.79 & 8.51   & 17.89 \\
 & 72   & 17.34 & 8.28   & 17.51 \\
 & 96   & 17.05 & 8.12   & 17.23 \\
 & 120  & 16.86 & 8.02   & 17.05 \\
 & 150  & 16.50 & 7.83   &  16.74 \\
 \bottomrule
\end{tabular}
\caption{Comparison between results using original annotations and annotations adapted to the live scenario for \textit{ActivityNet Captions}, employing LVC as the model.}
\label{tab:evaluation_gt_mod}
\end{table}

In conclusion, adapting annotations to how live models work has a positive impact on their evaluation.
However, the proposed adaptation has the drawback that the annotation must be dynamically adjusted to the temporal window $\Delta t$ being used, to then employ traditional offline metrics.
Furthermore, it is not an evaluation strategy that naturally provides a metric that allows us to observe the temporal evolution of the model.
In other words, these offline metrics do not evolve over time with the video, a fundamental aspect for the novel LVC problem.
All these drawbacks are clearly addressed by the new metrics proposed in this work.

We conclude that the analysis performed in this section is crucial to justify the need for a new evaluation metric for live models, so that we can evaluate them efficiently and fairly, as we shown in the following section.

\subsection{Evaluating LVC models with the novel Live Score}

One of the main motivations of our work has been to design a new online evaluation metric for live video captioning models, i.e., the LiveScore (LS) (see Section \ref{sec:tls}).
In this section, we detail all the experimental evaluations carried out using it.
Note that we provide a detailed experimental analysis considering the four alternatives proposed for the LS metric:

\begin{itemize}
    \item Normal operation mode (\textit{Live Score} - LS).
    \item Mode with a correction factor (\textit{Weighted Live Score} - wLS).
    \item Mode with history in a memory window (\textit{LS with History Window} - hLS).
    \item Combined mode with correction factor and history in a memory window (\textit{Weighted LS with History Window} - whLS).
\end{itemize}

In Table \ref{table:online_metrics}, we show the mean obtained by the novel LS when integrated with the different scorers Bleu4, METEOR, and ROUGE\_L.
Based on the results obtained, we can draw the following conclusions.
First, it is observed that the metrics increase as the parameter $\Delta t$ increases.
This makes sense, as the larger the temporal window, the LVC model is able to see more portions of the video and offer a better description in the caption.
However, values higher than $\Delta t = 120$ do not seem to offer a significant improvement.
Second, the results reported by the online metric are considerably higher than those obtained by the offline metrics.
This becomes clear when comparing the results between Table \ref{table:evaluation_gt_orig} and Table \ref{table:online_metrics}, where the results with the offline and online metrics are shown, respectively.
For example, for $\Delta t = 120$, the Bleu4 scorer goes from $1.55$ to $19.45$ for the LS version, or to $18.96$ for the wLS version.
The increase experienced for the rest of the scorers is somewhat similar.
This demonstrates the suitability of the new metric for the online scenario.
Third, when comparing between the LS and wLS versions, we observe how LS offers higher results than the wLS version.
This is because false positives that are generated are not considered by LS, but only by the wLS version, thus offering lower scores, but more adjusted to the actual performance of the LVC model.
Fourth, given the results offered by the different scorers and the LS and wLS metrics, the LVC model achieves the best results for $\Delta t = 120$,  with consensus among all scorers.

\begin{table}[t]
\centering
\begin{tabular}{@{}ll|cccccc@{}}
\toprule
 & \multirow{2}{*}{\textbf{Scorer}} & \multicolumn{6}{c}{LVC - $\mathbf{\Delta t}$}   \\ \cmidrule(lr){3-8}
&  & 24 & 48 & 72 & 96 & 120 & 150  \\ \midrule
\multirow{3}{*}{\textbf{LS}}
          & Bleu4   & 18.13 & 19.00 & 19.29 & 19.38 & \textbf{19.45} &
19.25  \\
          & METEOR  &  8.98 &  9.32 & 9.37  & 9.38  & \textbf{9.40} & 9.25   \\
          & ROUGE\_L& 19.84 & 20.81 & 21.14 & 21.21 & \textbf{21.29} & 21.18  \\
 \midrule
\multirow{3}{*}{\textbf{wLS}}
          & Bleu4   & 17.12 &  18.19  & 18.61 & 18.80 & \textbf{18.96}&
18.84  \\
          & METEOR  &  8.45 &  8.91 & 9.03  & 9.09  & \textbf{9.16} & 9.05   \\
          & ROUGE\_L& 18.71 & 19.90 & 20.36 & 20.55 & \textbf{20.74}& 20.72  \\
\bottomrule
\end{tabular}
\caption{Evaluation for the LVC model using the new online metric. LS: Live Score. wLS: Weighted LS. We evaluate the LVC model for different values of the parameter $\Delta t$, as well as integrating different scorers (Bleu4, METEOR, and ROUGE\_L).}
\label{table:online_metrics}
\end{table}

Finally, we include the analysis of the temporal evolution of the performance of our LVC model using all our novel evaluation metrics.
This analysis demonstrates that the new metrics enable us to generate graphs that track the temporal evolution of different scores.
Such tracking is crucial for LVC models and cannot be achieved with traditional offline metrics.
Note that we include the hLS and whLS variations of the metric in this analysis, which are specifically designed to continuously visualize the accuracy of LVC systems.

\begin{figure*}[ht!]
    \centering
    \includegraphics[trim={0 53.2cm 0 0},clip,width=0.98\linewidth]{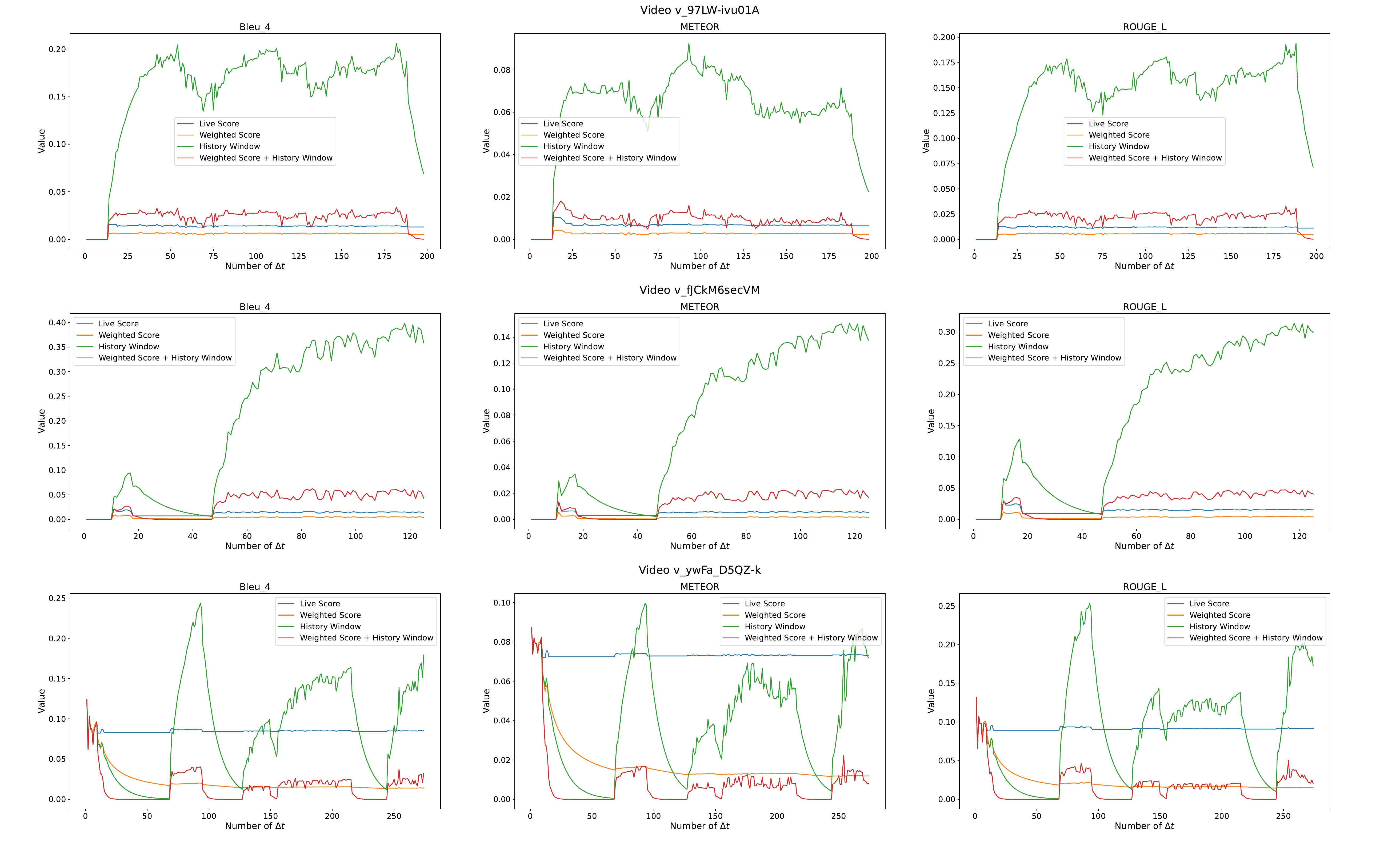}
    \caption{Temporal evolution of the designed online metrics for the LVC model with $\Delta t = 24$. Results are shown for 3 videos from ActivityNet Captions.}
    \label{fig:deltat_24}
\end{figure*}

\begin{figure*}[ht!]
    \centering
    \includegraphics[trim={0 53.2cm 0 0},clip,width=0.98\linewidth]{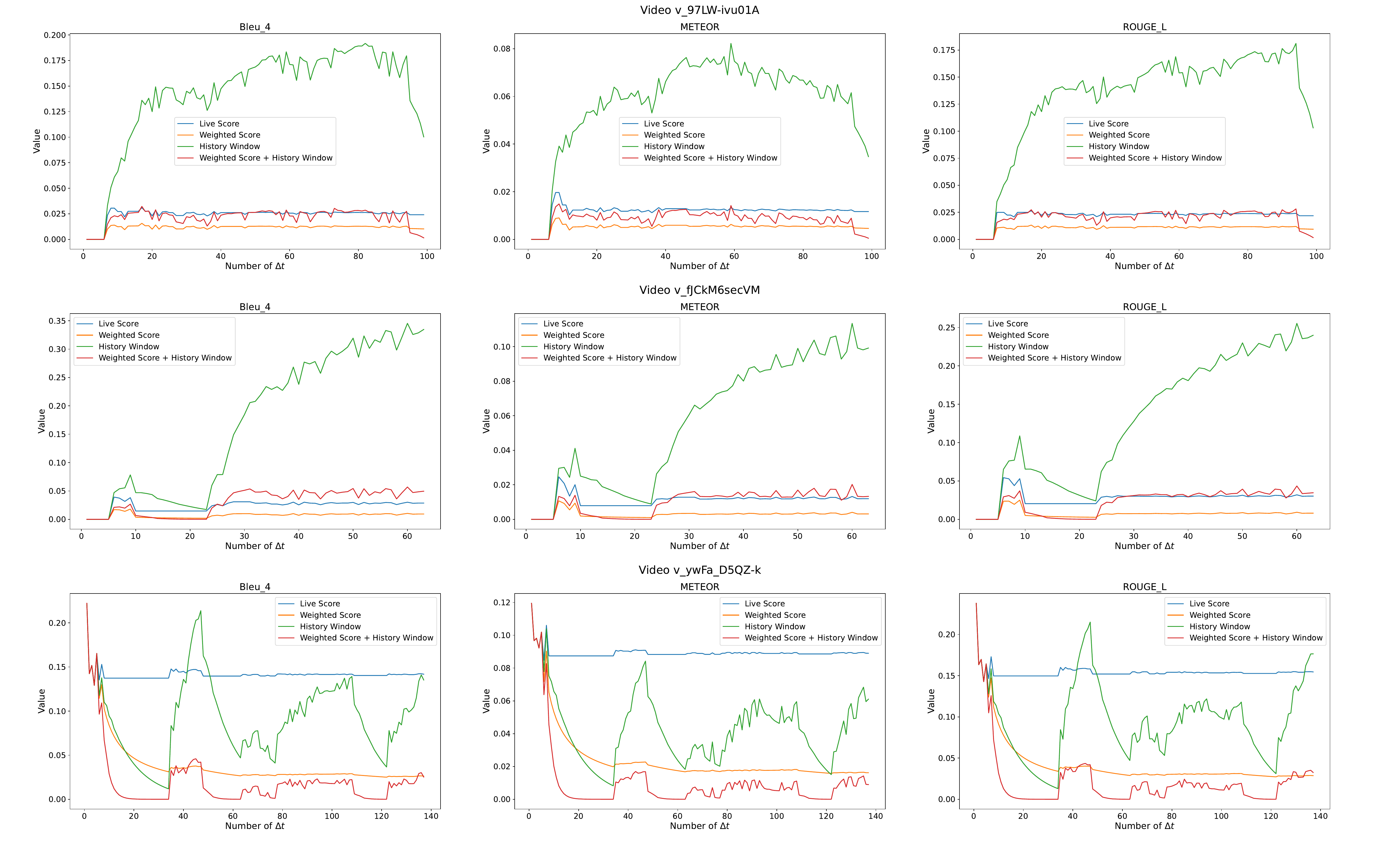}
    \caption{Temporal evolution of the designed online metrics for the LVC model with $\Delta t = 48$. Results are shown for 3 videos from ActivityNet Captions.}
    \label{fig:deltat_48}
\end{figure*}

\begin{figure*}[ht!]
    \centering
    \includegraphics[trim={0 53.2cm 0 0},clip,width=0.98\linewidth]{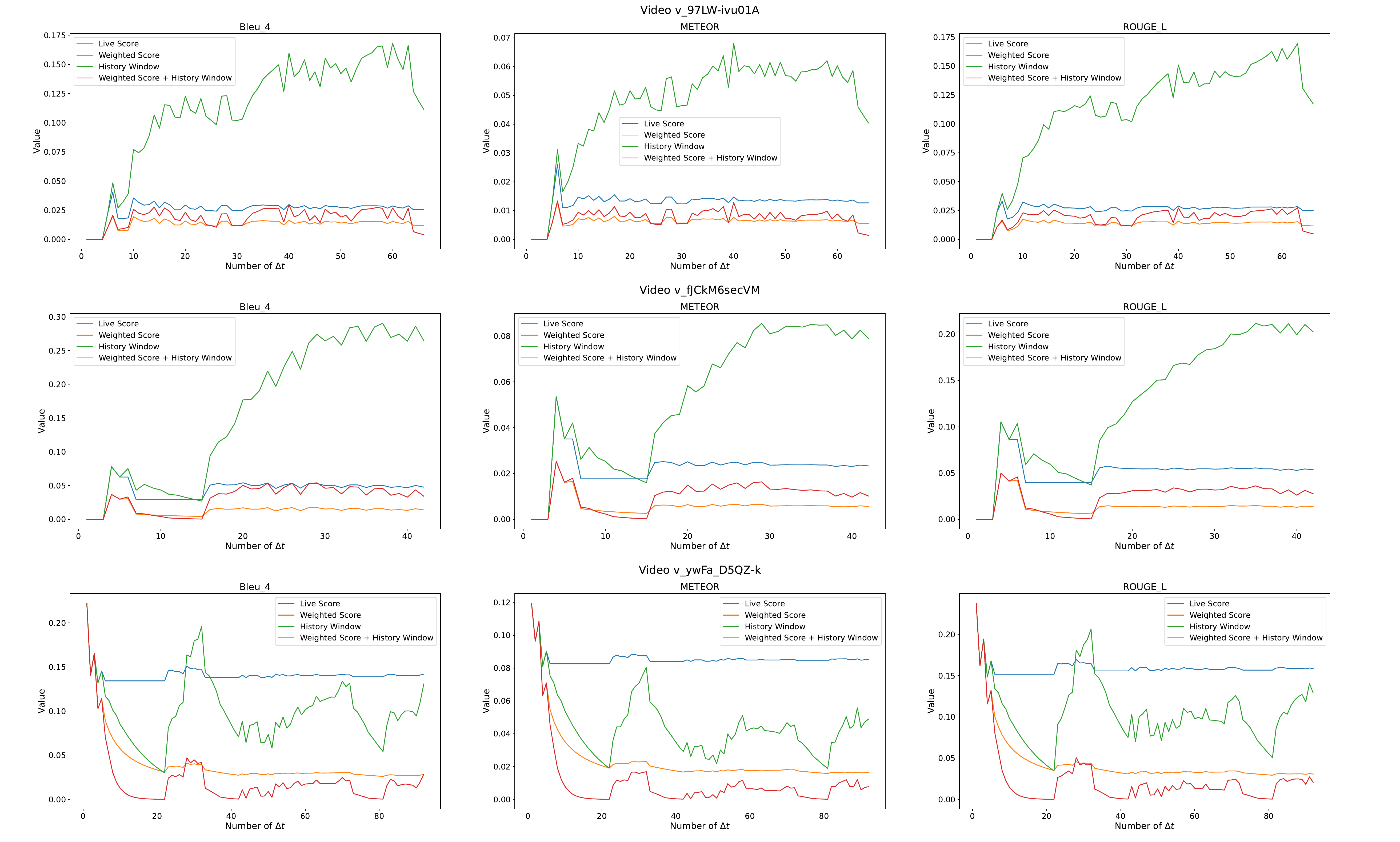}
    \caption{Temporal evolution of the designed online metrics for the LVC model with $\Delta t = 72$. Results are shown for 6 videos from ActivityNet Captions.}
    \label{fig:deltat_72}
\end{figure*}

\begin{figure*}[ht!]
    \centering
    \includegraphics[trim={0 53.2cm 0 0},clip,width=0.98\linewidth]{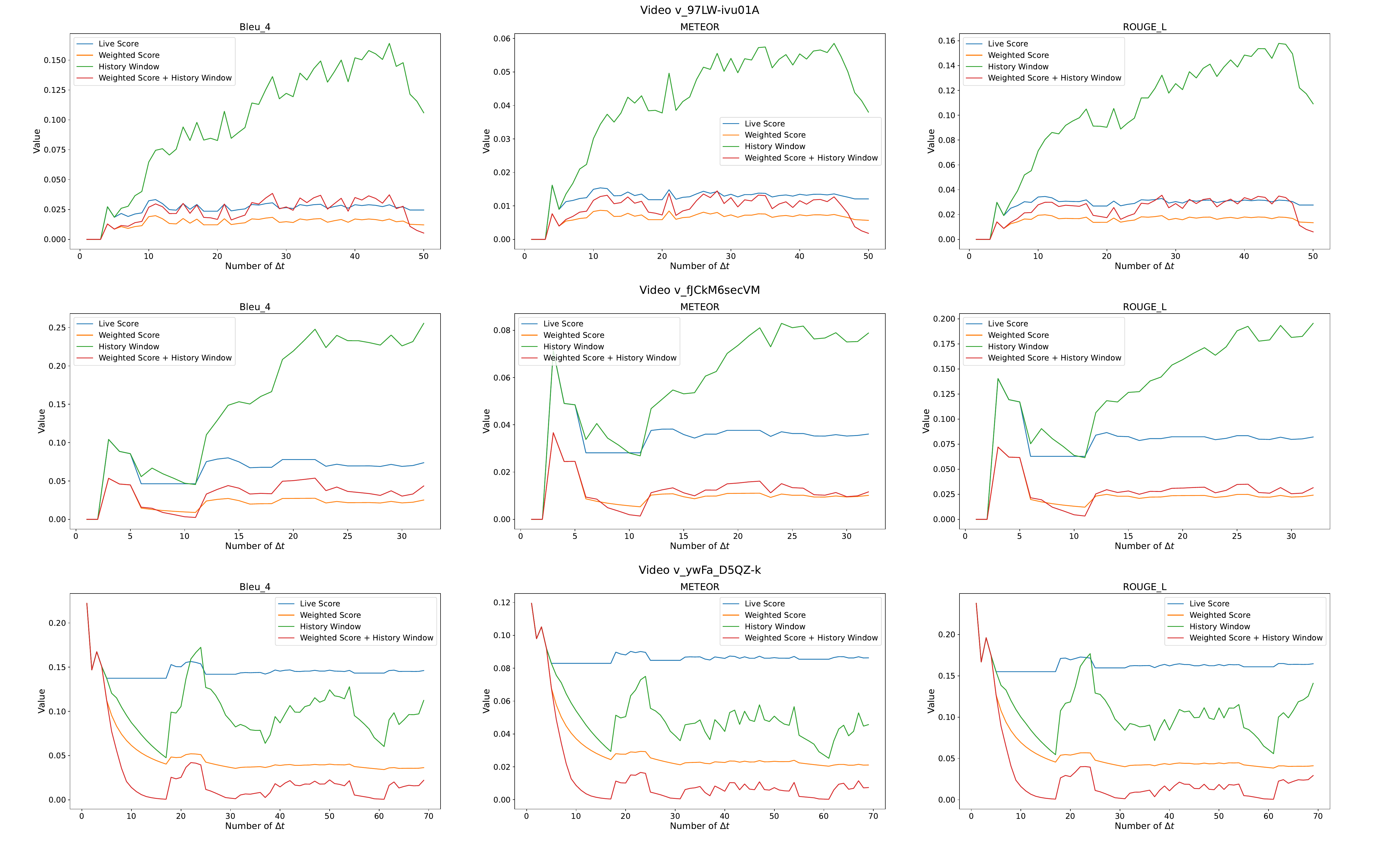}
    \caption{Temporal evolution of the designed online metrics for the LVC model with $\Delta t = 96$. Results are shown for 3 videos from ActivityNet Captions.}
    \label{fig:deltat_96}
\end{figure*}

\begin{figure*}[ht!]
    \centering
    \includegraphics[trim={0 53.2cm 0 0},clip,width=0.98\linewidth]{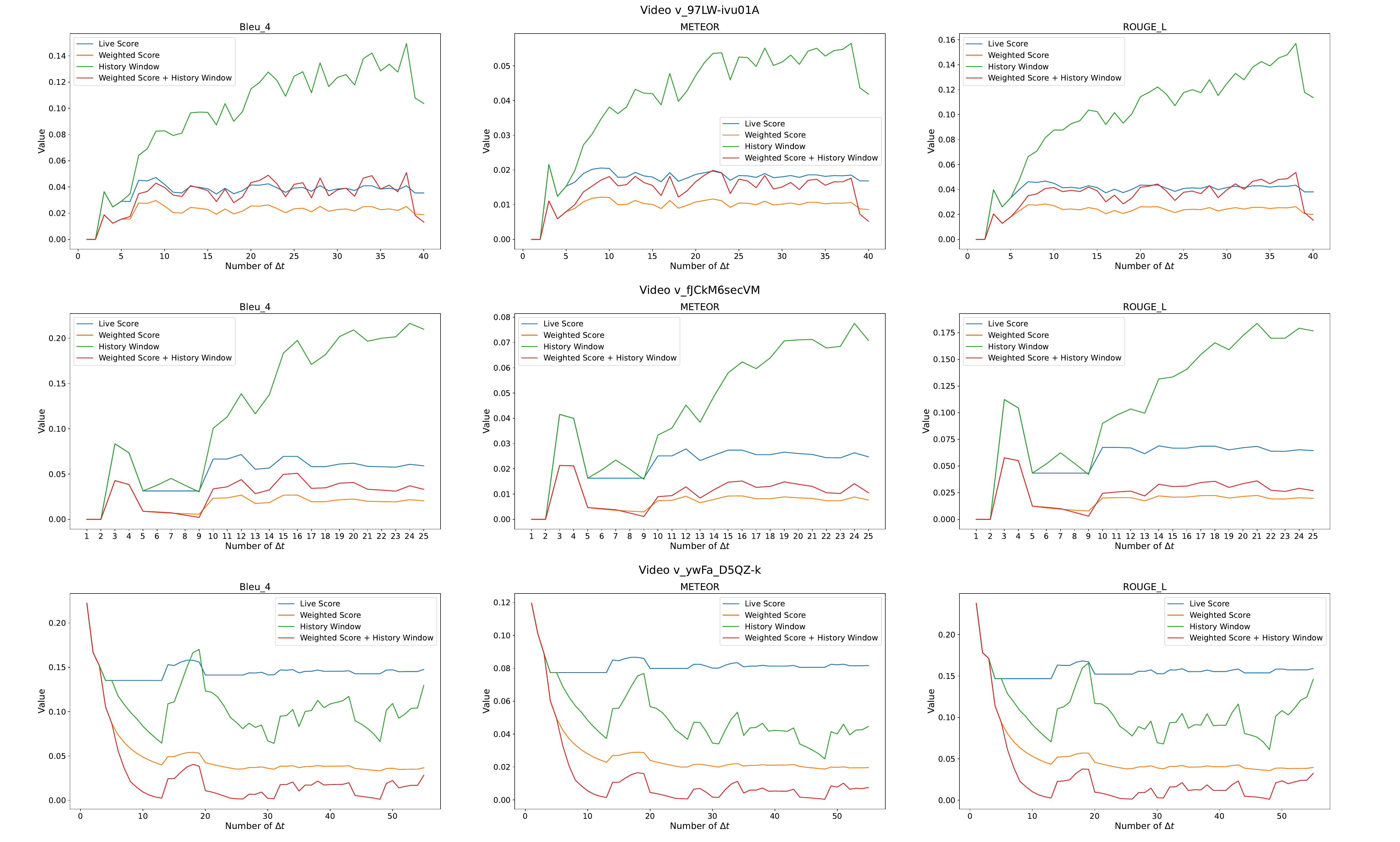}
    \caption{Temporal evolution of the designed online metrics for the LVC model with $\Delta t = 120$. Results are shown for 3 videos from ActivityNet Captions.}
    \label{fig:deltat_120}
\end{figure*}

\begin{figure*}[ht!]
    \centering
    \includegraphics[trim={0 53.2cm 0 0},clip,width=0.98\linewidth]{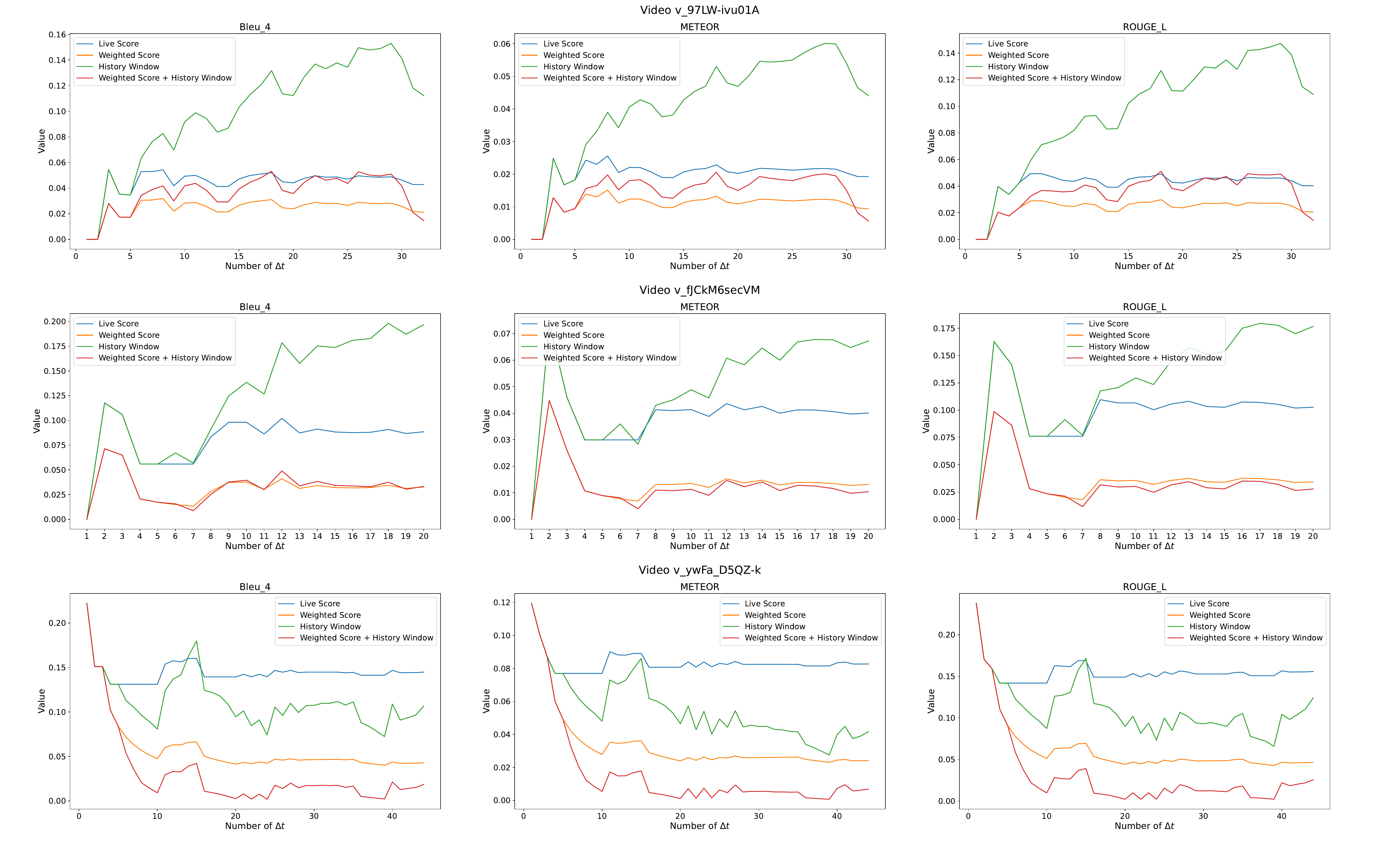}
    \caption{Temporal evolution of the designed online metrics for the LVC model with $\Delta t = 150$. Results are shown for 3 videos from ActivityNet Captions.}
    \label{fig:deltat_150}
\end{figure*}

In Figures \ref{fig:deltat_24} to \ref{fig:deltat_150}, we show the results obtained for 3 videos and all the values of the parameter $\Delta t$ used.
Analyzing these graphs, we can draw the following conclusions.
Firstly, the granularity of the metrics naturally increases as we decrease the parameter $\Delta t$.
This allows us to control the speed at which captions are generated and the speed at which their accuracy is evaluated.
Secondly, we have also included LS and wLS metrics without the temporal window option in these graphs.
We can observe how for all videos and all scorers, the wLS version is always more conservative, reporting lower or equal scores, as it applies a penalty based on false positives.
The same behavior is observed when comparing hLS and whLS, with the latter offering the most conservative scores.
It is also noticeable that all versions start by reporting exactly the same values, beginning to diverge when the history window comes into play.
As the third conclusion, perhaps the most important one, these plots show how the inclusion of a history window in our metric allows the model to exhibit its recovery in terms of caption generation accuracy.
It no longer considers the entire past, only the recent local past, so that the scorers can increase or recover as the LVC model chains more and more correct caption predictions.
In fact, observing a downward trend in the hLS and whLS metrics is an important indicator of how poorly the caption generator system is performing.


\subsubsection{Discussion}

While novel LS metric provides a robust online evaluation framework for LVC systems, it is not used directly during their training, as it can be observed in previous experiments.
LS calculates a cumulative score over time, capturing the real-time and historical accuracy of predictions, which contrasts with the frame-by-frame optimization process used during training of our LVC model.
Training objectives typically rely on instantaneous feedback, that are easier to optimize with stochastic gradient descent. Exploring methods to incorporate LS principles into a training objective remains an intriguing avenue for future research, potentially bridging the gap between evaluation and model optimization.

Another interesting aspect to analyze in the context of LVC is the quality of the annotations in the training datasets.
The robustness of video captioning models heavily depends on the quality and diversity of annotations in the training datasets.
High-quality annotations ensure accurate temporal boundaries and rich semantic details, enabling the model to learn fine-grained associations between actions and descriptions. Conversely, noisy or inconsistent annotations can introduce ambiguity, reducing the model’s ability to capture temporal precision and action differentiation. Similarly, a lack of variety in annotated events or scenarios limits the model's capacity to generalize to unseen contexts, particularly in real-world applications with diverse and dynamic content. These factors are especially critical in LVC, where models must operate with partial observations and strict timing constraints. Addressing these limitations requires the development of datasets with comprehensive, high-quality annotations tailored to the online scenario.

\subsubsection{Ablation Studies}
\label{sec:ablation}

\begin{figure}[ht!]
    \centering
    \includegraphics[width=0.6\linewidth]{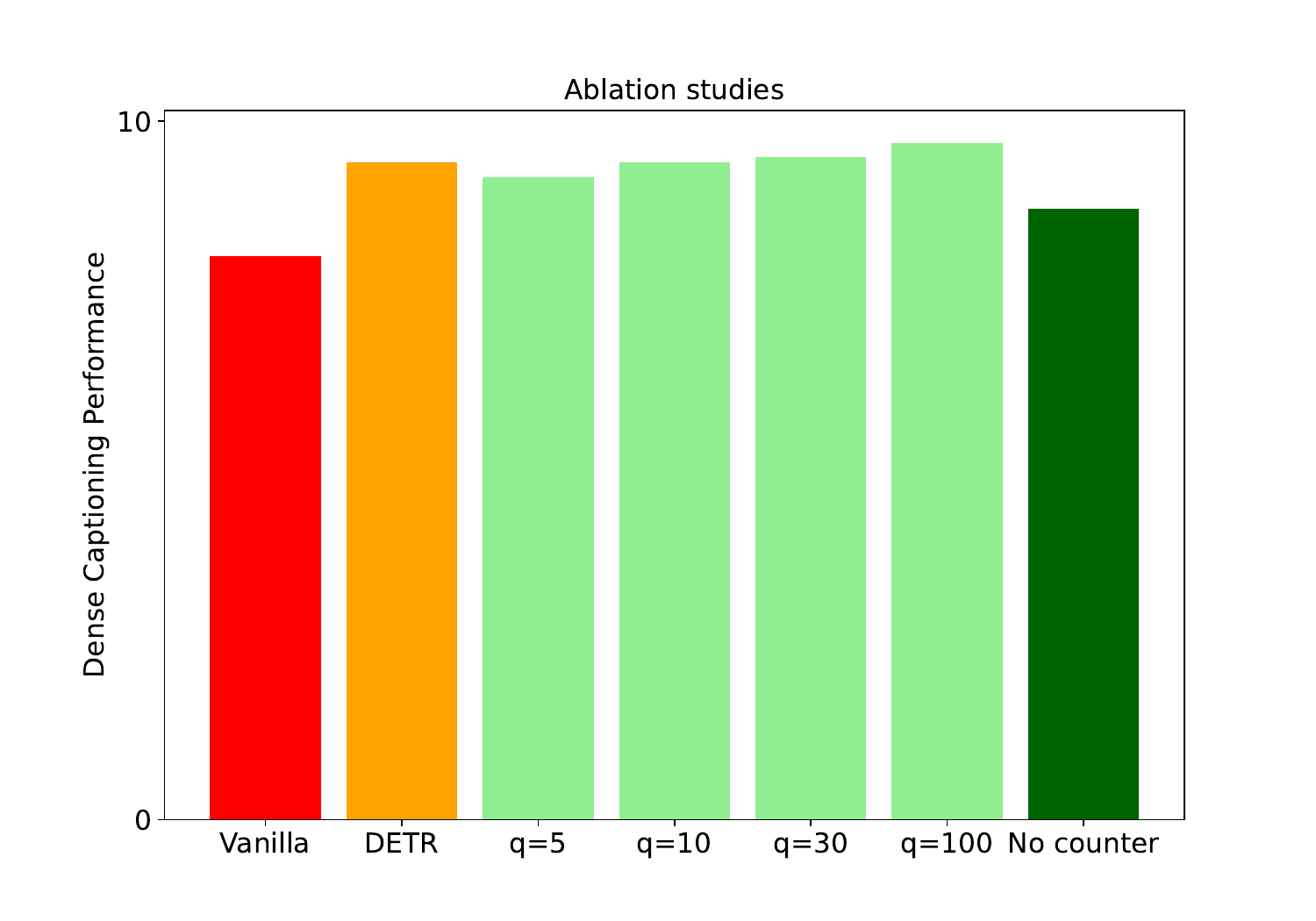}
    \caption{We show the impact of the different parameters and components of our LVC model on the quality of the dense captions when using a standard transformer (Vanilla) vs. a DETR, different values for $q$, and when the event counter head is not activated. We report METEOR with the LS metric.}
    \label{fig:ablation}
\end{figure}

In this section, we show the impact of the parameters of our LVC model on the quality of the dense captions.
Specifically, we analyze the influence of the following components of our model: the use of a deformable transformer vs. a standard transformer; the number of queries; and the event counter head.
For these studies we have used the METEOR metric in conjunction with our LS.
Figure \ref{fig:ablation} shows the results obtained.
As can be observed, the use of a deformable transformer significantly improves the results.
Regarding the number of queries used in our transformer, the impact is not as substantial, with $q=10$ being a fairly efficient compromise. Lastly, it is notable that not using the event counter head leads to a significant drop in the model's performance.

In summary, regarding the architecture used in the proposed LVC model, the use of a deformable transformer is confirmed as the option that provides the best results. The influence of the deformable transformer’s specific parameters, such as the number of queries, does not seem to significantly affect the quality of the captions. We believe this is because our model, instead of processing the entire video, works with short segments of the video, due to the online nature of the problem posed by LVC. Finally, the use of the three simultaneous heads—one for captions, another for localization, and another for considering the estimation of the number of events—appears to deliver the best results.

\subsection{Qualitative analysis}
\label{sec:qualitative_results}

In this section, we present some qualitative results of our LVC system.
Specifically, we show two examples corresponding to two videos from the validation set: one demonstrating a good result and the other a poor result.
We used $\Delta t = 150$, the maximum value we experimented with.
This makes it easier to identify points in the graph where there is a change in slope according to the proposed metrics.
We demonstrate the operation of the metric using the \textit{Live Score} strategy, integrating the \textit{Bleu4} metric.
In both examples, we highlight several points on the graph where notable changes in slope occur.
A pronounced change in slope indicates a significant improvement or deterioration in the score compared to previous points.
These marked points of interest represent where the predictions either align more closely or diverge from the ground truth.
Red shading signifies a greater discrepancy between the prediction and the ground truth, while green shading indicates a closer match.

\begin{figure}[t]
    \centering
    \includegraphics[width=\linewidth]{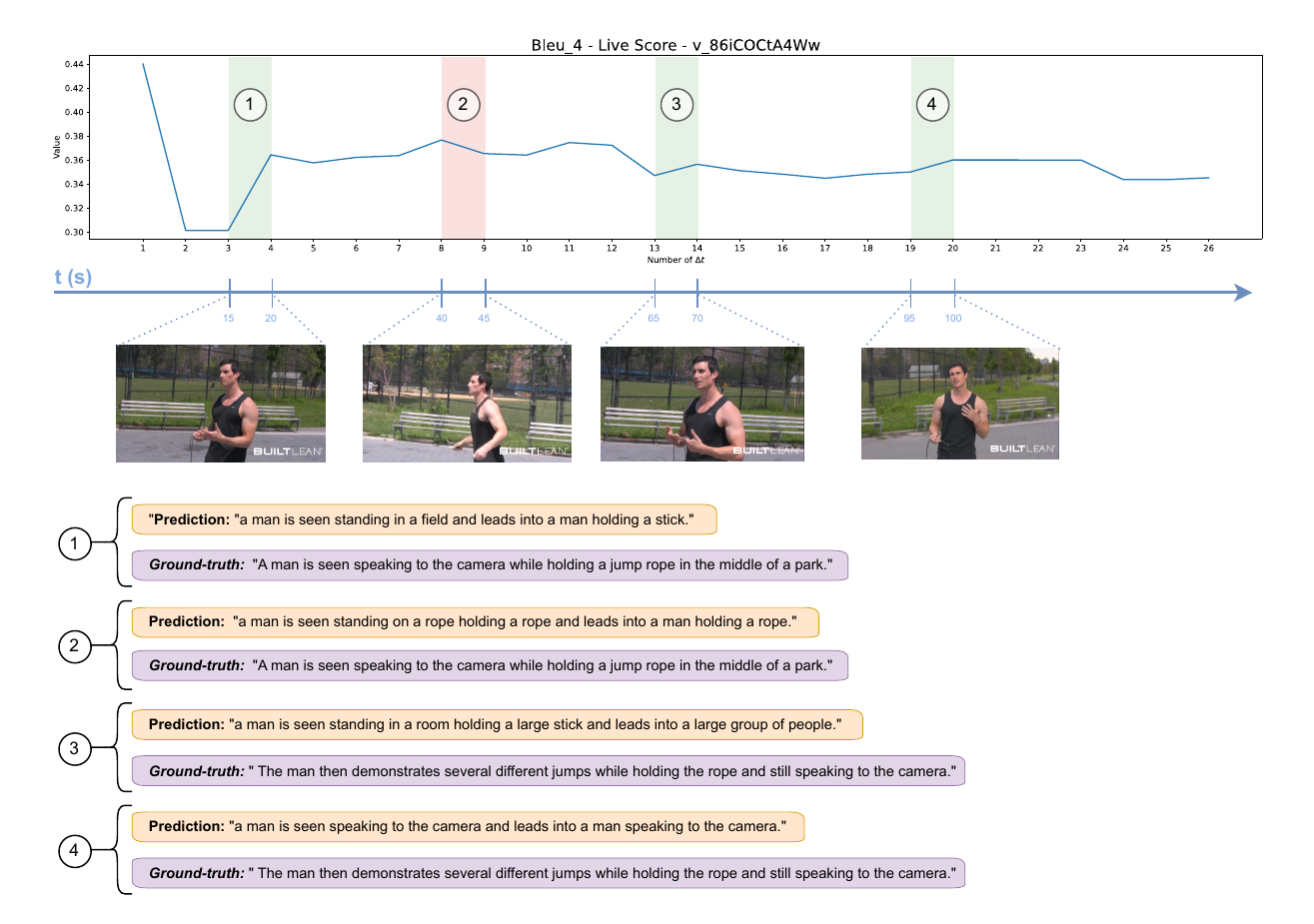}
    \caption{\textbf{Good Quality Example}: It is observed that the predictions of the LVC system resemble those that are annotated.}
    \label{fig:qualitative_good}
\end{figure}

\begin{figure}[t]
    \centering
    \includegraphics[width=\linewidth]{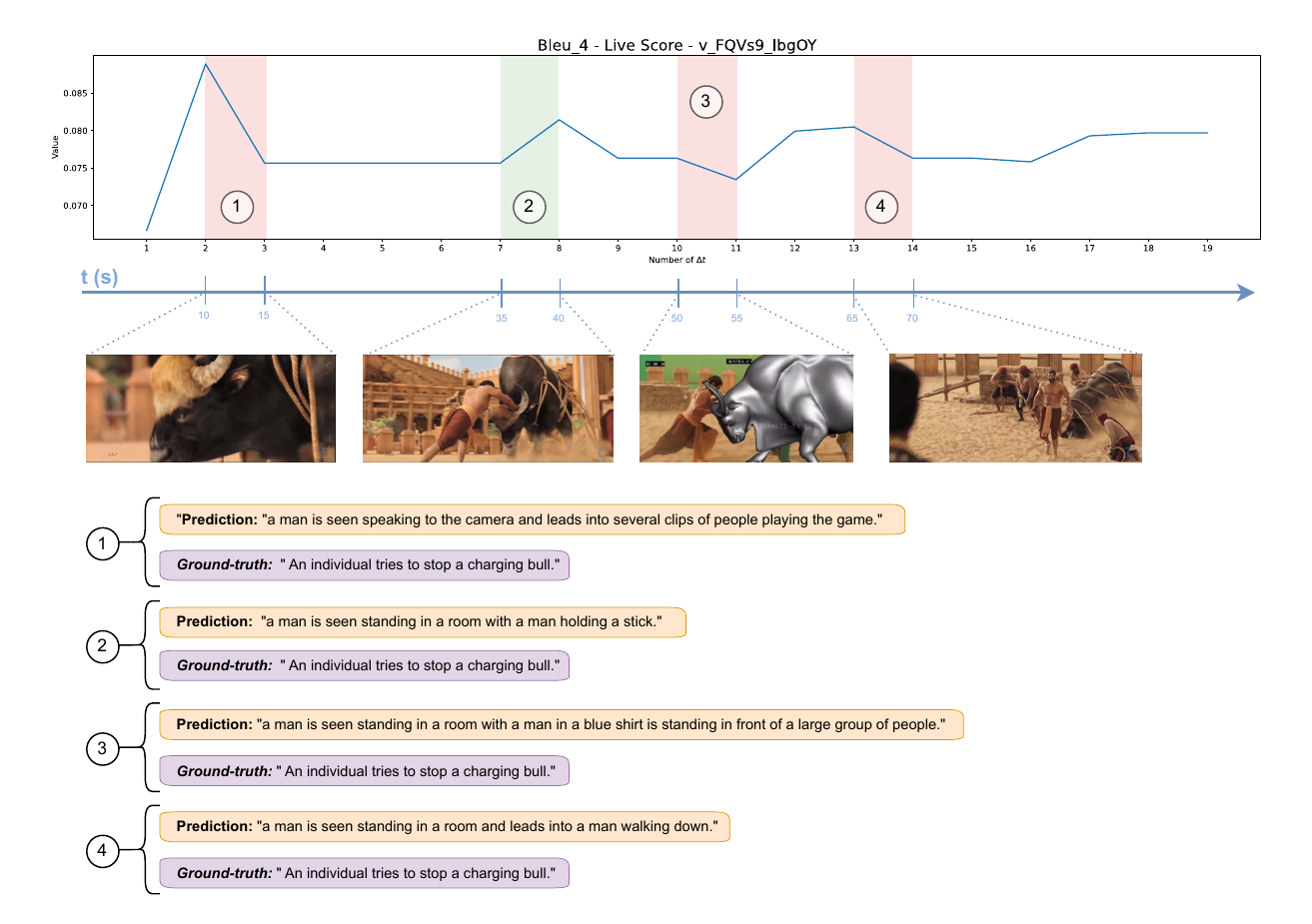}
    \caption{\textbf{Bad Quality Example}: The predictions do not resemble the annotations in the video, and the proposed metric reflects the system's failed behavior in an online mode.}
    \label{fig:qualitative_bad}
\end{figure}

Figure~\ref{fig:qualitative_good} contains the good case.
We can observe that it represents a scenario where the predictions closely resemble the ground-truth.
The scores obtained by the metric are quite good, in fact.
Figure~\ref{fig:qualitative_bad} represents a case where the predictions differ significantly from the ground-truth and the scores are low.

\subsection{Demo}\label{sec:demo}

In this section, we showcase the system in operation with a live video stream and highlight its performance.
We have implemented a demo that allows for the direct processing of a video stream from a camera using our LVC model.

\begin{figure}[t]
    \centering
    \includegraphics[width=0.8\linewidth]{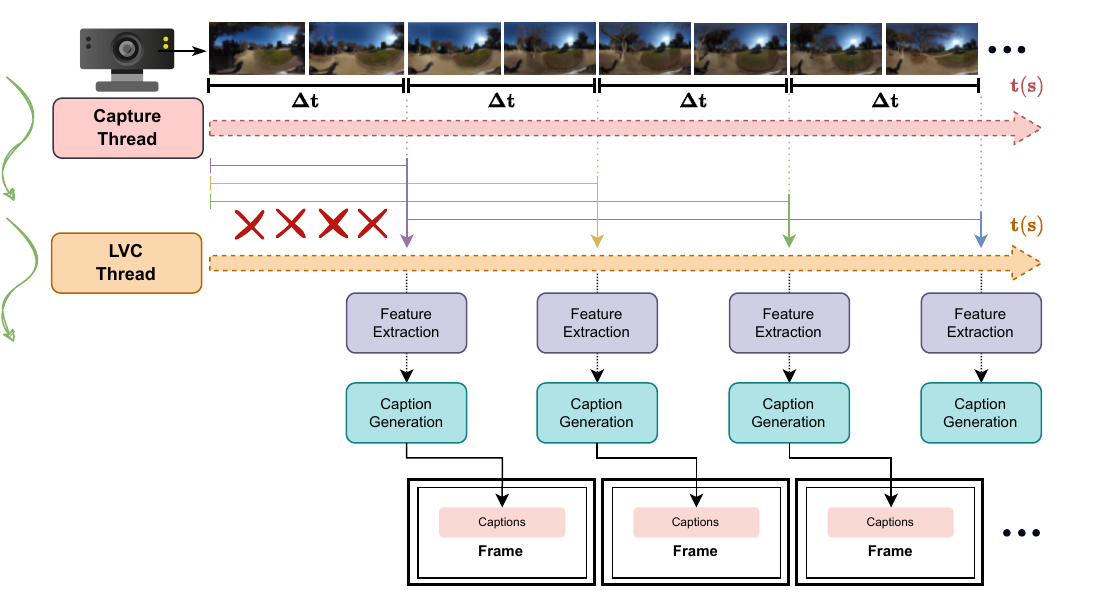}
    \caption{Multiprocessing demonstration scheme implemented. Our solution is capable of continuously displaying and producing captions. The system depicted in this figure employs a memory parameter $M=3$, so that the latest caption generation only receives the 2 previous segments and the current one.}
    \label{fig:demo_implementation}
\end{figure}

To achieve processing speed and provide captions with good quality, we have had to make the following interventions in the LVC model.
The demonstrator works by directly accessing a video stream from a webcam.
Once launched, the model will begin generating captions on the video stream immediately and continuously, displaying them on the screen.
We have implemented a multiprocessing solution with two threads running simultaneously.
The multiprocessing implementation is essential to ensure that while a caption is being generated, frames continue to be captured to avoid losing information.
The first thread is responsible for capturing frames and displaying the images and captions on the screen.
The second thread processes sets of frames to produce the captions using our LVC model.
This caption generation thread can be parameterized to define both the parameter $\Delta t$, i.e., the length of the minimum video segment to be analyzed, and a memory parameter $M$ that is maximum number of segments with length $\Delta t$ that we will keep in memory to produce captions in the demonstrator.
With this memory, we ensure that the dense caption prediction LVC module can access more context, a larger portion of the video, without losing its essence as a live system, and generate more accurate captions.
Figure \ref{fig:demo_implementation} provides a graphical description of the implementation made for the demonstrator.

As for processing speed, in Table~\ref{tab:results_demo}, we report the average frames per second that our
implementation is capable of processing and the average time it takes to generate a caption prediction by our LVC implementation.
The camera we are working with provides a frame rate of 30 frames per second, so we can calculate the optimal length of the video segments to be used by the LVC system.
We define $l$ as the length in frames of the input video clip: $l = 2.38 \cdot 30 \approx 71$
If the video segments we introduce to the model have a duration of 71 frames, we will ensure that when the processing
of one clip ends, the next one is introduced into the LCV model.
This way, no frames are discarded.
All these tests for the demo were performed on a laptop running Ubuntu 18.04 operating system, equipped with an Intel Core i7 processor, and an integrated NVIDIA Quadro RTX 5000 MaxQ graphics card.
Note that higher speed can be achieved if a more powerful GPU is used.

\begin{table}[ht!]
\centering
\scalebox{0.9}{
\begin{tabular}{@{}l|cc@{}}
\toprule
 &  \textbf{FPS} interfaz & \textbf{Time of prediction (s)}\\
\midrule
\textbf{Average}   & 18.93  & 2.38 \\
\bottomrule
\end{tabular}
}
\caption{Number of frames per second (FPS) and average prediction time reported by our LVC demo system.}
\label{tab:results_demo}
\end{table}

For the LVC problem, it is crucial to assess the performance of our model on devices with limited computational capacity, such as those used in embedded systems, robotics, IoT, or Edge Computing. To this end, we evaluated our LVC model on an NVIDIA Jetson Xavier-NX board, a platform equipped with 384-core NVIDIA Volta GPUs and 48 Tensor Cores, capable of operating with a power consumption as low as 20W. This setup allowed us to analyze the model's suitability for resource-constrained environments.
To conduct these performance tests, we adapted the developed demo to operate efficiently on the NVIDIA Jetson Xavier-NX.
First, we compiled the necessary libraries specifically for optimal use of the ARM64 architecture and the integrated GPU of the Jetson Xavier-NX board.
After embedding the application into the platform, we analyzed the system's performance, adjusting the parameter $\Delta t$ as shown in Table \ref{table:jetson_performance}.
One key observation is that our implementation achieves a frame rate exceeding 0.8 FPS, which could be sufficient for many applications requiring an embedded LVC system.
Notably, the maximum $\Delta t$ value that the Jetson platform can process is 70 frames.
It is worth highlighting that the processing of TSP features represents the main bottleneck.
In percentage terms, the time spent generating these features consistently exceeds the time required for generating dense captions and filtering when $\Delta t > 16$ frames.

\begin{table}[ht!]
\centering
\scalebox{0.9}{
\begin{tabular}{@{}l|ccc@{}}
\toprule
 $\Delta t$&  \textbf{FPS} & \textbf{Time of features} (sec.) & \textbf{Time of LVC} (sec.)\\
\midrule
16   & 0.87  & 0.39  & 0.75 \\
48   & 0.51  & 1.15  & 0.75 \\
70   & 0.11  & 7.6  & 2.4 \\
\bottomrule
\end{tabular}
}
\caption{Performance evaluation of our LVC model on an NVIDIA Jetson Xavier-NX board. We report: number of frames per second (FPS), and average time for the compuation of features and captions.}
\label{table:jetson_performance}
\end{table}

\section{Conclusions}

Live video captioning is a novel and challenging problem that has not been deeply investigated in the scientific
literature.
As we have shown, generating dense captions for live video streams is a much harder problem than one might conclude
from results reported in previous works under more constrained settings, e.g., offline dense video captioning models.
In fact, traditional evaluation metrics need to be updated to novel online versions that allow us to judge the actual
live performance of the LVC models.
In this work, we have formalized the problem of LVC for the first time, proposed new metrics tailored to it.
We introduced an LVC model capable of integrating transformer-based attention mechanisms with a caption filtering module for video streams received as input.
Results of our model as well as an evaluation kit with the novel metrics integrated are made publicly
available to encourage further research on LVC on realistic data: \url{https://github.com/gramuah/lvc}.
We hope to encourage more researchers to look into the challenging yet very practical task of LVC.
This work enhances our understanding of LVC and paves the way for live video understanding and accessibility applications in dynamic environments.

As future work, we consider the following options.
The first would be to allow our LVC model to improve the predictions of the captions, using a memory mechanism, where the model takes into account past predictions.
Technically, for example, recurrent networks could be incorporated into the model.
A second line of improvement would consist of incorporating explainability techniques to improve the interpretability of our LVC model. Inspired by advancements in Explainable Artificial Intelligence (XAI) \citep{Akkem2024}, such as attention visualization and feature importance analysis, we aim to provide insights into the model’s decision-making process. These methods could help bridge the gap between black-box models and transparent systems, enhancing trust and usability in applications requiring streaming video analysis.
Finally, the future capabilities LVC systems are closely tied to advancements in sensor technology and video analysis techniques. Improvements in camera sensors, such as higher frame rates, and low-light performance, can provide richer data for more accurate caption generation. Additionally, emerging video analysis techniques, including multimodal feature fusion and lightweight neural architectures, may enhance the efficiency and robustness of LVC systems, particularly for real-time applications. While these developments are beyond the scope of this study, they represent promising directions for expanding the applicability and effectiveness of LVC models in diverse scenarios, including robotics, surveillance, and assistive technologies.

\backmatter

\section*{Declarations}

\begin{itemize}
    \item Acknowledgements: We appreciate the assistance of Walfrido González Molina in generating some figures.
    \item Funding: This research was partially funded by projects: NAVISOCIAL, with reference 2023/00405/001 from the University of Alcal\'a; NAVIGATOR-D, with reference PID2023-148310OB-I00 from the Ministry of Science and Innovation of Spain.
    \item Conflict of interest/Competing interests: The authors certify that they have no conflict of interest.
    \item Ethics approval and consent to participate: Not applicable.
    \item Consent for publication: Not applicable.
    \item Data availability: The dataset used for all the experiments in this paper is \textit{ActivityNet Captions}~\citep{krishna2017dense}, which can be accessed at \url{https://cs.stanford.edu/people/ranjaykrishna/densevid/}.
    \item Materials availability: Not applicable.
    \item Code availability: Code to reproduce our results can be found at \url{https://github.com/gramuah/lvc}.
    \item Author contribution: Conceptualization, all; methodology, R.J.L.-S. and E. B.-F.;
software, R.J.L.-S., E. B.-F., N.N. and C.G.-Á.; validation, all; ; writing—original draft preparation, all; supervision, R.J.L.-S. and S. M.-B.; project administration, R.J.L.-S. and S. M.-B.; funding acquisition, R.J.L.-S. and S. M.-B..
\end{itemize}

\bibliography{library}
\end{document}